\documentclass[10pt,twocolumn,letterpaper]{article}
\usepackage[pagenumbers]{cvpr}
\usepackage{graphicx}
\usepackage{amsmath}
\usepackage{amssymb}
\usepackage{booktabs}
\usepackage{fontawesome}
\usepackage{wrapfig}
\usepackage{xspace}
\usepackage[dvipsnames]{xcolor}
\definecolor{Urlcolor}{RGB}{251,111,146}
\definecolor{Linkcolor}{RGB}{193,18,31}
\definecolor{CiteColor}{RGB}{32,126,190}
\usepackage[pagebackref,breaklinks,colorlinks,urlcolor=Urlcolor,linkcolor=Linkcolor,citecolor=CiteColor]{hyperref}
\usepackage[accsupp]{axessibility}
\usepackage[capitalize]{cleveref}
\crefname{section}{Sec.}{Secs.}
\Crefname{section}{Section}{Sections}
\Crefname{table}{Table}{Tables}
\crefname{table}{Tab.}{Tabs.}
\Crefname{equation}{Equation}{Equations}
\crefname{equation}{eq.}{eqs.}

\newcommand{\tinyspace}{\kern 0.025em}
\newcommand{\stemp}{\faChild\@\xspace}
\newcommand{\skey}{\faKey\@\xspace}
\newcommand{\sview}{\faCamera\@\xspace}
\newcommand{\smask}{\faScissors\@\xspace}
\newcommand{\svid}{\faVideoCamera\@\xspace}
\newcommand{\sflow}{\faForward\@\xspace}

\newcommand{\supcmr}{\sview\tinyspace\skey\tinyspace\smask}
\newcommand{\supucmr}{\stemp\tinyspace\smask}
\newcommand{\supumr}{\smask}

\newcommand{\supacsm}{\stemp\tinyspace\smask}
\newcommand{\supdove}{\smask\tinyspace\sflow\svid}
\newcommand{\supours}{\smask}

\newcommand{\method}{MagicPony\xspace}

\makeatletter
\renewcommand{\paragraph}{%
\@startsection{paragraph}{4}%
{\z@}{0.5em}{-1em}%
{\normalfont\normalsize\bfseries}%
}
\makeatother

\newcommand\rurl[1]{%
  \href{https://#1}{\nolinkurl{#1}}%
}

\newif\ifarxiv
\arxivtrue

\def\cvprPaperID{409}
\def\confName{CVPR}
\def\confYear{2023}

\title{\method: Learning Articulated 3D Animals in the Wild}
\author{Shangzhe Wu\thanks{Equal contribution.}
\quad
Ruining Li$^*$
\quad
Tomas Jakab$^*$
\quad
Christian Rupprecht
\quad
Andrea Vedaldi \\[0.3em]
Visual Geometry Group, University of Oxford\\
{\tt\small \{szwu, ruining, tomj, chrisr, vedaldi\}@robots.ox.ac.uk}\\[0.1em]
\small\rurl{3dmagicpony.github.io}
}

\begin{document}
\twocolumn[\maketitle\vspace{-3em}\begin{center}
    \includegraphics[trim={0 0 30px 0}, clip, width=0.99\linewidth]{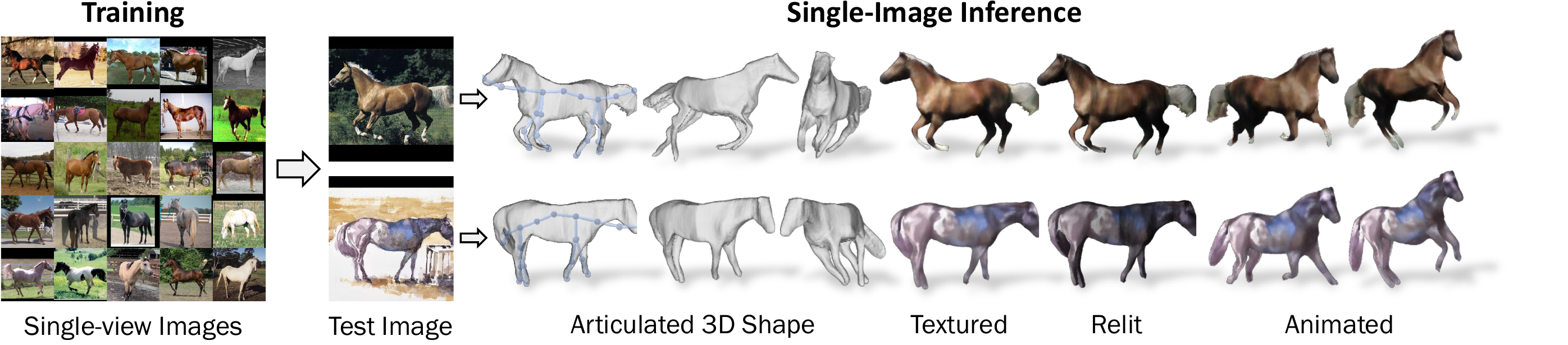}
\end{center}\vspace{-1.5em}
\captionof{figure}{\textbf{Learning Articulated 3D Animals in the Wild.} Our method trains on a collection of single-view images of an animal category and produces a model that can predict the articulated 3D shape of a new instance from a single test image, which can be animated and relit.
}%
\label{fig:teaser}\bigbreak]
\def\thefootnote{*}\footnotetext{Equal contribution.}\def\thefootnote{\arabic{footnote}}

\begin{abstract}
We consider the problem of predicting the 3D shape, articulation, viewpoint, texture, and lighting of an articulated animal like a horse given a single test image as input.
We present a new method, dubbed \method, that learns this predictor purely from in-the-wild single-view images of the object category, with minimal assumptions about the topology of deformation.
At its core is an implicit-explicit representation of articulated shape and appearance, combining the strengths of neural fields and meshes.
In order to help the model understand an object's shape and pose, we distil the knowledge captured by an off-the-shelf self-supervised vision transformer and fuse it into the 3D model.
To overcome local optima in viewpoint estimation, we further introduce a new viewpoint sampling scheme that comes at no additional training cost.
\method outperforms prior work on this challenging task and demonstrates excellent generalisation in reconstructing art, despite the fact that it is only trained on real images.
The code can be found on the project page at {\small\url{https://3dmagicpony.github.io/}}.
\end{abstract}

\section{Introduction}%
\label{s:intro}

Reconstructing the 3D shape of an object from a single image of it requires knowing \emph{a priori} what are the possible shapes and appearances of the object.
Learning such a prior usually requires ad-hoc data acquisition setups~\cite{Anguelov05, Liu13, loper2015smpl, Joo18}, involving at least multiple cameras, and often laser scanners, domes and other hardware, not to mention significant manual effort.
This is viable for certain types of objects such as humans that are of particular interest in applications, but it is unlikely to scale to the long tail of objects that can appear in natural images.
The alternative is to learn a 3D prior from 2D images only, which are available in abundance.
However, this prior is highly complex and must be learned \emph{while using it to reconstruct in 3D} the 2D training data, which is a major challenge.

In this paper, we propose \textbf{\method}, a novel approach to learning 3D models of articulated object categories such as horses and birds with only \emph{single-view} input images for training.
We leverage recent progress in unsupervised representation learning, unsupervised image matching, efficient implicit-explicit shape representations and neural rendering, and devise a new auto-encoder architecture that reconstructs the 3D shape, articulation and texture of each object instance from a single image.
For training, we only require a 2D segmenter for the object category and a description of the topology and symmetry of its 3D skeleton (\ie, the number and connectivity of bones).
We \emph{do not} require apriori knowledge of the objects' 3D shapes, keypoints, viewpoints, or of any other 2D or 3D cue which are often used in prior work~\cite{Kar15, kanazawa18cmr, Goel20ucmr, li20online}.
From this, we learn a function that, at test time, can estimate the shape and texture of a new object from a single image, in a feed-forward manner.
The function exhibits remarkable generalisation properties, including reconstructing objects in \emph{abstract drawings}, despite being trained on real images only.

In order to learn such disentangled 3D representations simply from raw images, \method addresses a few key challenges.
The first challenge is viewpoint estimation.
Before the 3D model is available, it is very difficult to assign a viewpoint to the objects in the training images.
To tackle this problem, prior works have often assumed additional cues such as 2D keypoint correspondences~\cite{Kar15, kanazawa18cmr, li20online}.
Instead, we avoid requiring additional keypoint supervision and implicitly infer noisy correspondences by \emph{fusing} into the 3D model knowledge distilled from DINO-ViT~\cite{caron2021dino}, a self-supervised visual transformer network (ViT)~\cite{Kolesnikov21vit}.
We also develop a new efficient disambiguation scheme that explores multiple viewpoint assignment hypotheses at essentially no cost, avoiding local optima that are caused by greedily matching the noisy 2D correspondences.

The second challenge is how to represent the 3D shape, appearance and deformations of the object.
Most prior works have used textured meshes~\cite{kanazawa18cmr, Goel20ucmr, li20online, Li2020umr, wu2021dove, yang21lasr, goel2022differentiable}, but these are difficult to optimise from scratch, leading to problems that often require ad-hoc heuristics such as re-meshing~\cite{yang21lasr, goel2022differentiable}.
The other and increasingly popular approach is to use a volumetric representation such as a neural radiance field~\cite{mildenhall2020nerf, reizenstein2021common, barron2022mipnerf, verbin2022refnerf}, which can model complex shapes, including manipulating their topology during training.
However, this modelling freedom comes at the cost of over-parametrisation, which is particularly problematic in monocular reconstruction and often leads to meaningless short-cut solutions~\cite{verbin2022refnerf}.
Furthermore, modelling articulation with a volumetric representation is difficult.
A posing transformation is only defined for the object's surface and interior and is more easily expressed from the canonical/pose-free space to the posed space.
However, rendering a radiance field requires transforming 3D points off the object surface and in the direction opposite to the posing transformation, which is hard~\cite{chen2021snarf}.

We address these issues by using a hybrid volumetric-mesh representation based on DMTet~\cite{shen2021dmtet, munkberg2021nvdiffrec}.
Shape and appearance are defined volumetrically in canonical space, but a mesh is extracted on the fly for posing and rendering.
This sidesteps the challenges of using neural rendering directly while retaining most of its advantages and enabling the use of powerful shape regularisers.

To summarise, we make the following \textbf{contributions}:
(1) A new 3D object learning framework that combines recent advances in unsupervised learning, 3D representations and neural rendering, achieving better reconstruction results with less supervision;
(2) An effective mechanism for fusing self-supervised features from DINO-ViT into the 3D model as a form of self-supervision;
and (3) an efficient multi-hypothesis viewpoint prediction scheme that avoids local optima in reconstruction with no additional cost.

We compare our method to prior work on several challenging articulated animal categories and show that our method obtains \emph{significantly better} quantitative and qualitative results while using \emph{significantly less} supervision, and also demonstrate generalisation to abstract drawings.

\section{Related Work}
\begin{table} [t!]
    \setlength{\tabcolsep}{1.2pt}
    \footnotesize
	\centering
    \caption{\textbf{Related Work Overview on Weakly-supervised Learning of 3D Objects.} Annotations: \stemp template shape, \sview viewpoint, \skey 2D keypoint, \smask object mask, \sflow optical flow, \svid video, $^1$coarse template shape from keypoints, $^2$camera estimated from keypoints using SfM, $^3$outputs texture flow, $^4$shape bases initialised from CMR\@. $^\dagger$UMR relies on part segmentations from SCOPS~\cite{hung19scops}.
    }
	\begin{tabular}{lcccccccccccc}
		\toprule
		& \multicolumn{6}{c}{Supervision} & \multicolumn{5}{c}{Output} \\
		\cmidrule(lr){2-7} \cmidrule(lr){8-12}
		Method                                          &    \stemp         &  \sview          & \skey      &  \smask     & \sflow      & \svid         & \vspace{1.5pt} 3D            & 2.5D          & Motion        & View          & Texture           \\
        \midrule
		Unsup3D~\cite{wu20unsupervised}                 &                   &                  &            &             &             &               &               & \checkmark    &               &  \checkmark   & \checkmark        \\
		CSM~\cite{kulkarni19canonical}                  &  \checkmark       &                  &            & \checkmark  &             &               &               &               &               & \checkmark    &                   \\
		A-CSM~\cite{kulkarni20articulation-aware}        &  \checkmark       &                  &            & \checkmark  &             &               &               &               & \checkmark    & \checkmark    &                   \\
        CMR~\cite{kanazawa18cmr}                   & (\checkmark)$^1$  & (\checkmark)$^2$ & \checkmark & \checkmark  &             &               & \checkmark    &               &               & \checkmark    & (\checkmark)$^3$  \\
        U-CMR~\cite{Goel20ucmr}                        &  \checkmark       &                  &            & \checkmark  &             &               & \checkmark    &               &               & \checkmark    & \checkmark        \\
        UMR$^\dagger$~\cite{Li2020umr}        &                   &                  &            & \checkmark  &             &               & \checkmark    &               &               & \checkmark    & (\checkmark)$^3$  \\
        ACMR~\cite{li20online}                           &  (\checkmark)$^4$ & (\checkmark)$^2$ & \checkmark & \checkmark  &             &               & \checkmark    &               &               & \checkmark    & (\checkmark)$^3$  \\
        DOVE~\cite{wu2021dove}                          &                   &                  &            &  \checkmark & \checkmark  & \checkmark    & \checkmark    &               & \checkmark    & \checkmark    & \checkmark        \\
        \midrule
        Ours                                            &                   &                  &            &  \checkmark &             &               & \checkmark    &               & \checkmark    & \checkmark    & \checkmark        \\
		\bottomrule
	\end{tabular}\label{table:related}
    \vspace{-0.1in}
\end{table}

\paragraph{Weakly-supervised Learning of 3D Objects.}
Most related to our work are weakly-supervised methods for single-image 3D object reconstruction that learn from a collection of images or videos of an object category~\cite{kato2018renderer, kanazawa18cmr, liu19soft, kato2019learning, wang2018pixel2mesh, henderson2019learning, Goel20ucmr, Li2020umr, li20online, wu2021derender, kokkinos2021point, Kokkinos_2021_CVPR, wu2021dove}.
Due to the inherent ambiguity of the problem, these methods usually rely on heavy geometric supervision (see \cref{table:related}) in addition to object masks, such as 2D keypoints~\cite{kanazawa18cmr, li20online}, template shapes~\cite{Goel20ucmr, kulkarni19canonical, kulkarni20articulation-aware}.
UMR~\cite{Li2020umr} forgoes the requirement of keypoints and template shapes by leveraging weakly-supervised object part segmentations (SCOPS~\cite{hung19scops}),
but produces coarse symmetric shapes similar to CMR~\cite{kanazawa18cmr}.
DOVE~\cite{wu2021dove} learns coarse articulated objects by exploiting temporal information from videos with optical flow supervision.
All these methods require object masks for supervision,
except Unsup3D~\cite{wu20unsupervised} which exploits bilateral symmetry for reconstructing roughly frontal objects, like faces,
and UNICORN~\cite{monnier2022unicorn} which uses a progressive conditioning strategy with a heavily constrained bottleneck, leading to coarse reconstructions.
Another emerging paradigm is to leverage generative models~\cite{nguyen2019hologan, Schwarz2020graf, chan2021piGAN, poole2022dreamfusion} that encourage images rendered from viewpoints sampled from a prior distribution to be realistic.
These models, however, often rely heavily on an accurate estimation of viewpoint distribution and/or a powerful 2D image generative model, both of which are difficult to obtain in small-data scenarios.

\paragraph{Optimization from Multi-views and Videos.}
3D reconstruction traditionally relies on epipolar geometry of multi-view images of static scenes~\cite{Faugeras1993, hartley04multiple}.
Neural Radiance Fields (NeRF)~\cite{mildenhall2020nerf, reizenstein2021common, barron2022mipnerf, verbin2022refnerf} have recently emerged as a powerful volumetric representation for multi-view reconstruction given accurate cameras.
A recent line of work, LASR~\cite{yang21lasr}, ViSER~\cite{yang2021viser} and BANMo~\cite{yang2022banmo}, optimises 3D shapes of deformable objects from a small set of monocular videos, with heavily engineered optimisation strategies using optical flow and mask supervision, and additionally DensePose~\cite{neverova20continuous} annotations for highly deformable animals.
Concurrent work of LASSIE~\cite{yao2022lassie} also leverages DINO-ViT~\cite{caron2021dino} image features for supervision but optimises a part-based model on a small collection of images ($\sim$30) of an object category.
It uses a shared shape model for all instances with only per-instance articulation.

\paragraph{Learnable Deformable 3D Representations.}
A common 3D representation for weakly-supervised shape learning is triangular meshes.
Deformation on meshes can be modelled by estimating offsets of individual vertices~\cite{kanazawa18cmr, li20online, wu2021dove}, which typically requires heavy regularisation, such as As-Rigid-As-Possible (ARAP)~\cite{sorkine2007rigid}.
To constrain the space of deformations, many works often utilise lower-dimensional models, such as cages~\cite{yifan2020neural} or linear blend skinning controlled with skeletal bones~\cite{loper2015smpl, wu2021dove} or Gaussian control points~\cite{yang21lasr, yang2021viser}.
Parametric models, like SMPL~\cite{loper2015smpl} and SMAL~\cite{zuffi20173d}, allow for realistic control of the shape through learned parameters but often require an extensive collection of 3D scans to learn from.

Dynamic neural fields~\cite{pumarola20d-nerf, park2021nerfies, gafni2020dynamic, tretschk2021nonrigid, raj2020pva, noguchi2021neural, su21a-nerf:, yang2022banmo}, such as D-NeRF~\cite{pumarola20d-nerf}, extend NeRF~\cite{mildenhall2020nerf} with time-varying components.
A-NeRF~\cite{su21a-nerf:} proposes a radiance field conditioned on an articulated skeleton, while BANMo~\cite{yang2022banmo} uses learned Gaussian control points similarly to~\cite{yang21lasr,yang2021viser}.
One key limitation of these implicit representations is the requirement of an inverse transformation from 3D world coordinates back to the canonical space, which is often harder to learn compared to forward deformation~\cite{chen2021snarf}.
We propose a hybrid SDF-mesh representation, extending DMTet~\cite{shen2021dmtet} with an articulation mechanism, which combines the expressiveness of implicit models with the simplicity of mesh deformation and articulation.

\paragraph{Viewpoint Prediction in the Wild.}
Viewpoint prediction in weakly-supervised settings is challenging, as it is prone to local optima induced by projection and common object symmetries.
Existing solutions are based on learning a distribution over multiple possible viewpoints.
U-CMR~\cite{Goel20ucmr} proposes a camera-multiplex that optimises over 40 cameras for each training sample using a given template shape, 
which is expensive as evaluating each hypothesis involves a rendering step.
We propose a new viewpoint exploration scheme that requires evaluating \emph{only} one hypothesis in each step with essentially no added training cost and does not rely on a given template shape but a jointly learned model.
Recent work of Implicit-PDF~\cite{implicitpdf2021} proposes to learn viewpoint distributions implicitly by estimating a probability for each image-pose pair of a fixed object instance.
RelPose~\cite{zhang2022relpose} extends this to learning relative poses of image pairs.
Our model also predicts a  probability associated with each viewpoint hypothesis but instead relates an instance to a learned category-level prior model.%

\section{Method}
\begin{figure*}[t]
\centering
\includegraphics[width=\linewidth]{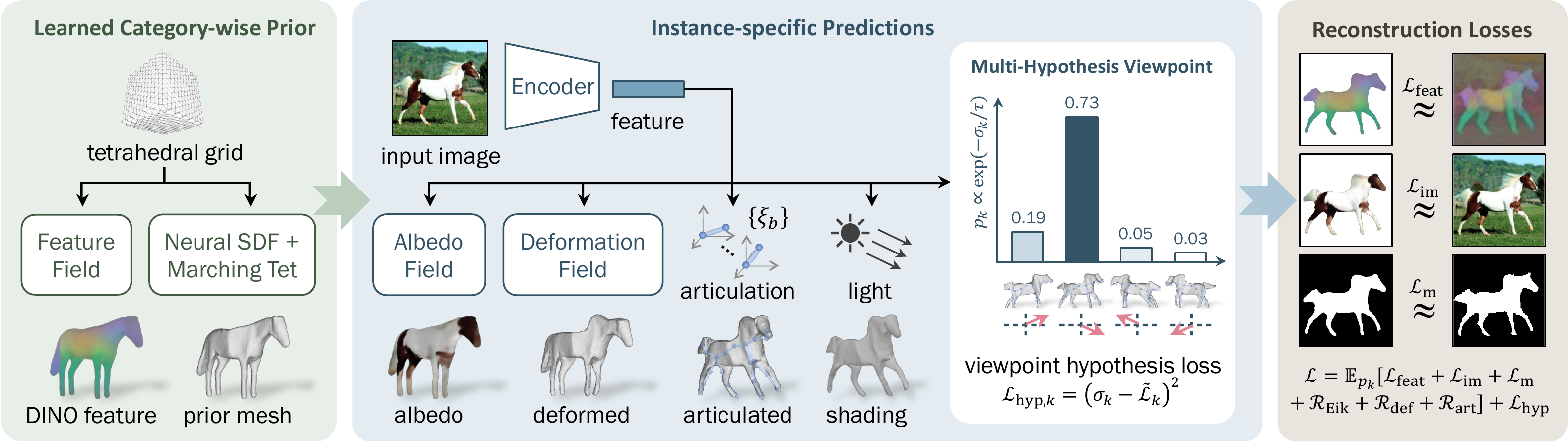}
\caption{\textbf{Training Pipeline.}
Given a collection of single-view images of an animal category, our model learns a category-specific prior shape using an implicit-explicit representation, together with a feature field that allows us to fuse self-supervised correspondences through a feature rendering loss.
This prior shape is deformed, articulated and shaded based on features extracted from a training image.
To combat local minima in viewpoint prediction, we introduce an efficient scheme that explores multiple hypotheses at essentially no additional cost.
The entire pipeline is trained with reconstruction losses end-to-end, except for the frozen image encoder pre-trained via self-supervision.}%
\label{fig:method}
\end{figure*}

\newcommand{\x}{\boldsymbol{x}}
\newcommand{\bu}{\boldsymbol{u}}

Given only a collection of single-view images of a deformable object category collected in the wild, our goal is to train a monocular reconstruction network that, at test time, predicts the 3D shape, articulated pose, albedo and lighting from a single image of the object.

During training, our method requires no geometric supervision other than the images, 2D foreground segmentations of the object obtained automatically from an off-the-shelf method like~\cite{kirillov2019pointrend}, and a description of the topology of the object's skeleton (\eg, the number of legs).
It is based on several key ideas: an implicit-explicit shape representation (\cref{s:implicit-explicit}), a hierarchical shape representation, from a generic template to articulated instances (\cref{s:articulated}), and a viewpoint prediction scheme that leverages self-supervised correspondences and efficient multi-hypothesis exploration to avoid local minima (\cref{s:viewpoint}), as illustrated in \cref{fig:method}.

\subsection{Implicit-Explicit 3D Shape Representation}%
\label{s:implicit-explicit}

The choice of representation for the 3D shape of the object is critical.
It must be (1) sufficiently \emph{expressive} for modelling fine-grained shape details and deformations and (2) sufficiently \emph{regular} for learning with only weak supervision via image reconstruction.

Most 3D representations are either explicit (triangular meshes~\cite{kanazawa18cmr, yang21lasr, wu2021dove}) or implicit (volumes, SDFs, radiance fields~\cite{occnet, Park2019DeepSDF, mildenhall2020nerf, wang2021neus, jang21codenerf:}).
Explicit representations are more compact, easily support powerful deformation models like blend skinning, and the geometric smoothness can be easily measured and controlled;
however, they are more difficult to optimise and often result in defects like folds during training, which requires occasional re-meshing~\cite{yang21lasr,goel2022differentiable}.
Implicit representations are more amenable to gradient-based optimisation, including allowing topological changes, but are more computationally expensive and, due to their high capacity, require substantial supervision, usually in the form of a dense coverage of the different object viewpoints~\cite{mildenhall2020nerf,wang2021neus} or 3D ground-truth~\cite{occnet, Park2019DeepSDF}.

Here we adopt a representation that combines the advantages of both.
The shape of the object is represented implicitly by a neural field that is converted on the fly into an explicit mesh via the \emph{Differentiable Marching Tetrahedra} (DMTet)~\cite{shen2021dmtet,munkberg2021nvdiffrec,Doi1991AnEM} method.
Specifically, the object's shape
$
S = \{\x \in \mathbb{R}^3 \vert s(\x) = 0\}
$
is defined as the isosurface of the signed distance function (SDF)
$
s: \mathbb{R}^3 \mapsto \mathbb{R}
$,
which is implemented by a Multi-Layer Perceptron (MLP) taking 3D coordinates as input.
To encourage $s$ to be a signed distance function, we sample random points $\x \in \mathcal{X}\subset\mathbb{R}^3$ in space and minimise the Eikonal regulariser~\cite{icml2020_2086}:
\begin{equation}
\mathcal{R}_\text{Eik}(s)
=
\frac{1}{|\mathcal{X}|}
\sum_{\x \in \mathcal{X}} (\|\nabla s(\x)\|_2 - 1)^2.
\end{equation}
We then evaluate the function $s$ at the vertices of a tetrahedral grid and extract a watertight mesh from it using DMTet~\cite{shen2021dmtet}, which is fast and differentiable.

We use this implicit-explicit representation to learn a category-wise template shape shared across all instances, and model instance deformations through a hierarchical pipeline, explained next.
The function $s$ is initialised with a generic ellipsoid elongated along the $z$-axis.
We denote by $V_\text{pr} \subset \mathbb{R}^3$ the vertices of the extracted prior mesh.

\subsection{Articulated Shape Modelling and Prediction}%
\label{s:articulated}

The category-wise template $V_\text{pr}$ is deformed into a specific object instance $V_\text{ins}$, and further posed to match the shape $V$ in each input image (\cref{fig:method}).
These deformations are predicted from a single image, as discussed below.

\paragraph{Image Encoder.}

The input image is first encoded by means of DINO-ViT~\cite{caron2021dino}, a self-supervised Vision Transformer (ViT)~\cite{Kolesnikov21vit}.
The ViT extracts from the image a set of key and output tokens, which were found to code, respectively, for correspondences~\cite{amir2021deep} and appearance~\cite{tumanyan2022splicing}.
Hence, we use the keys to predict the object's deformation and the outputs to predict its appearance.
Specifically, we apply the ViT $\Phi$ to the input image
$
I \in \mathbb{R}^{3 \times H \times W}
$
and extract the \emph{key} $\Phi_\text{k} \in \mathbb{R}^{D \times H_p \times W_p}$ and \emph{output} $\Phi_\text{o} \in \mathbb{R}^{D \times H_p \times W_p}$ \emph{tensors} from the last attention layer.
These features are local, with one token per image patch.
We use two small convolutional networks to further extract from these a global key
$
\phi_\text{k} = f_\text{k} (\Phi_\text{k}) \in \mathbb{R}^{D}
$
and output
$
\phi_\text{o} = f_\text{o} (\Phi_\text{o}) \in \mathbb{R}^{D}
$
vector.

\paragraph{Instance-specific Shape.}

The shape of each object instance $V_\text{ins}$ differs slightly from the template shape $V_\text{pr}$.
This is accounted for by the deformation
$
V_{\text{ins},i} = V_{\text{pr},i} + \Delta V_i,
$
where the displacement
$
\Delta V_i = f_{\Delta V} (V_{\text{pr},i}, \phi_\text{k})
$
of each vertex $V_{\text{pr},i}$ is predicted by an MLP $f_{\Delta V}$ from the global image features $\phi_\text{k}$.
We exploit the fact that many objects are bilaterally symmetric and enforce both the prior $V_\text{pr}$ and instance $V_\text{ins}$ shapes to be symmetric by mirroring the query locations for the MLPs about the $yz$-plane.
We also limit the magnitude of the the deformation $V_{\text{ins}}$ by adding the regularizer
$
\mathcal{R}_\text{def}(\Delta V) = \frac{1}{|V_\text{pr}|} \sum_i \| \Delta V_i \|_2^2.
$

\paragraph{Posed Shape.}

In order to match the 3D shape observed in the image, the instance-specific mesh $V_{\text{ins}}$ is further deformed by the \emph{posing function}
$
V_i = g(V_{\text{ins},i}, \xi)
$,
where $\xi$ denotes the pose parameters.
Posing accounts for the deformations caused by the articulation of the object's skeleton; they allow for large but controlled deformations which would otherwise be more difficult to account for by a generic deformation field like $\Delta V$.

We adopt the blend skinning model for posing~\cite{magnenat1988abstract,chadwick1989layered}, parameterised by $B-1$ bone rotations $\xi_b \in SO(3), b=2,\dots,B$, and the viewpoint $\xi_1 \in SE(3)$.
Specifically, we initialise a set of rest-pose joint locations $\mathbf{J}_b$ on the instance mesh using simple heuristics\footnote{A chain of bones going through the two farthest end points along $z$-axis for birds, and for quadrupeds, additionally four legs branching out from the body bone to the lowest point in each quadrant. See \ifarxiv \cref{sec:bone_topology}\else sup.~mat\fi.}.
Each bone $b$ but the root has one parent $\pi(b)$, thus forming a tree structure.
Each vertex $V_i$ is associated to the bones by the skinning weights $w_{ib}$, defined based on relative proximity to each bone (see \ifarxiv {}\cref{sec:skinning_weights}\else sup.~mat.\fi for details).
The vertices are then posed by the linear blend \emph{skinning equation}:
\begin{equation}\label{e:skinning}
\begin{aligned}
    V_i(\xi) =\left( \sum_{b=1}^B w_{ib} G_b(\xi) G_b(\xi^*)^{-1} \right) V_{\text{ins},i}, \\
    G_1 = g_1,
    ~~
    G_b = G_{\pi(b)} \circ g_b,
    ~~
    g_b(\xi) = \begin{bmatrix}
        R_{\xi_b} & \mathbf{J}_b \\ 0 & 1 \\
    \end{bmatrix},
\end{aligned}
\end{equation}
where $\xi^*$ denotes the bone rotations at the rest pose.

\paragraph{Pose Prediction.}

Similarly to the instance-specific deformation $\Delta V_i$, the bone rotations $\xi_{2:B}$ are predicted by a network $f_\text{b}$ from the image features $\Phi_\text{k}$ and $\phi_\text{k}$.
The viewpoint $\xi_1 \in SE(3)$ is predicted separately, as discussed in~\cref{s:viewpoint}.

Pose articulation depends on estimating local body joints and benefits from using local features.
To extract them, we first project the centroid of each bone at rest into the image plane using the predicted viewpoint $\xi_1$:
\begin{equation*}
    \bu_b = \Pi \left[ G_{\pi(b)}(\xi_1,\xi_{2:B}^*)  \frac{\mathbf{J}_{b}}{2} \right],
\end{equation*}
and sample the local features from the patch key feature map $\Phi_\text{k}(\bu_b)$ at the projected pixel locations $\bu_b$.

To predict bone rotations from the image features, we adopt a transformer to account for the inter-dependency of the bones, implied by the underlying skeleton.
Specifically, we use a small transformer network~\cite{transformer} $f_\text{b}$ that takes as input $B-1$ tokens
$
\nu_b = (\phi_\text{k}, \Phi_\text{k}(\bu_b), b, \mathbf{J}_b, \bu_b)
$
and outputs the bone rotations
$
(\xi_2, \ldots, \xi_B) = f_\text{b}(\nu_2, \ldots, \nu_B)
$
parameterised as Euler angles.
We also limit the deviation from the rest pose via the regulariser
$
\mathcal{R}_\text{art}(\xi) = \frac{1}{B-1} \sum_{b=2}^B \|\xi_b\|_2^2$.

\subsection{Appearance and Shading}%
\label{s:appearance}
We decompose the appearance of the objects into albedo and diffuse shading assuming a Lambertian illumination model.
The albedo of the object is modelled in the same ``canonical space'' as the shape SDF $s$ by a second coordinate MLP
$
a(\x) = f_\text{a}(\x, \phi_\text{o}) \in [0,1]^3
$,
which takes as input the global output features $\phi_\text{o}$.
The dominant light direction
$
l \in \mathbb{S}^2
$
and ambient and diffuse intensities
$
k_s, k_d \in \mathbb{R}
$
are predicted by a lighting network $f_\text{l}(\phi_\text{o})$.

The posed mesh is then rendered into an image via \emph{differentiable deferred mesh rendering}.
The renderer first rasterizes all pixels $\bu$ that overlap with the posed object $V$, computing the 3D coordinates $\x(\bu)$ of the corresponding object point in canonical space;
then the MLP $f_\text{a}$ is evaluated, once per pixel, to obtain the colour $a(\x(\bu))$.
The final pixel colour accounts for shading and is given by
$$
\hat I(\bu) = \left(k_s + k_d \cdot \max \{0, \langle l, n(\bu) \rangle \}\right) \cdot a(\x(\bu)),
$$
where $n(\bu) \in \mathbb{S}^2$ is the normal of the \emph{posed} mesh $V$.
Using the same differentiable renderer, we also obtain the predicted mask $\hat{M} \in [0,1]^{H\times W}$ of the object.

For training supervision, we use the appearance loss
$
\mathcal{L}_\text{im}(\hat I, I, M) = \| \tilde{M} \odot (\hat{I} - I) \|_1
$
as well as the mask loss
$
\mathcal{L}_\text{m}(\hat M, M) 
=
\| \hat{M} - M \|_2^2
+
\lambda_\text{dt}(\hat{M} \odot \texttt{dt}(M))
$,
where the distance transform $\texttt{dt}(\cdot)$ is used to obtain better gradients~\cite{kanazawa18cmr, wu2021derender}, $\tilde{M} = \hat{M} \odot M$ is the intersection of the rendered and ground-truth masks, and $\lambda_\text{dt}$ is a weighing factor.

\subsection{Viewpoint Prediction}%
\label{s:viewpoint}

Predicting the object viewpoint is a critical first step for learning 3D shapes but often difficult~\cite{kanazawa18cmr,Goel20ucmr}.
Here, we contribute a method to do so robustly and efficiently, which exploits self-supervised correspondences in a multi-hypothesis prediction pipeline, as detailed next.

\paragraph{Fusing Self-Supervised Correspondences.}

Estimating explicit correspondences without supervision is difficult and brittle;
hence, we propose to incorporate correspondence information implicitly and softly.
Inspired by N3FF~\cite{tschernezki22neural}, we \emph{fuse} information from the self-supervised ViT features into the 3D model by \emph{learning to render} them.%
\footnote{This is also similar to BANMo~\cite{yang2022banmo}, but we use self-supervised features instead of supervised DensePose~\cite{neverova20continuous} ones.}
This is done by adding a coordinate network
$
\psi(\x) \in \mathbb{R}^{D^{\prime}}
$
that predicts a $D^{\prime}$-dimensional feature vector for each 3D point $\x$ in canonical space.
This is similar to the albedo network $f_a$, except that it is shared across all instances of the same object category, and hence there is no dependency on the input image.
These canonical features are then projected into a feature image
$
\hat{\Phi}_\text{k} \in \mathbb{R}^{H \times W\times D'}
$
using the same differentiable renderer used for the albedo,
and trained to minimise a corresponding rendering loss:
\begin{equation}
\mathcal{L}_\text{feat}
(\hat{\Phi}_\text{k}, \Phi'_\text{k}, \tilde M)
=
\| \tilde{M} \odot (\hat{\Phi}_\text{k} - \Phi'_\text{k}) \|_2^2,
\end{equation}
where $\Phi'_\text{k} = \texttt{PCA}(\Phi_\text{k})$ is a PCA-reduced (to $D^{\prime}=16$ principal components) version of the original DINO-ViT features $\Phi_\text{k}$ for training efficiency.

\paragraph{Multi-Hypothesis Viewpoint Prediction.}

Prior works have found that a major challenge in learning the object's viewpoint is the existence of multiple local optima in the reconstruction objective.
Some have addressed this issue by sampling a large number of hypotheses for the viewpoint~\cite{Goel20ucmr}, but this is somewhat cumbersome and slow.
Instead, we propose a scheme that explores multiple viewpoints \emph{statistically}, but at each iteration samples only a single one, and thus comes ``for free''.

We hence task a viewpoint network $f_\text{vp}(\phi_k)$ to predict from $\phi_\text{k}$ four viewpoint rotation hypotheses $R_k \in SO(3), k \in \{1, 2, 3, 4\}$, each in one of the four quadrants\footnote{
For bilaterally symmetric animals, there are often four ambiguous orientations towards each of the four $xz$-quadrants in the canonical space.} around the object, as illustrated in \cref{fig:method}.
The network also predicts a score
$
\sigma_k
$
for each hypothesis, used to evaluate the probability $p_k$ that hypothesis $k$ is the best of the four options as
$
p_k \propto \exp(-\sigma_k/\tau)
$ using \texttt{softmax},\
where $\tau$ is a temperature parameter, gradually decreased during training.

The naive approach for learning $\sigma_k$ is to sample multiple hypotheses and compare their reconstruction loss to determine which one is better.
However, this is expensive as it requires rendering the model with all hypotheses.
Instead, we suggest to \emph{sample} a single hypothesis for each training iteration and simply train $\sigma_k$ to predict the expected reconstruction loss $\tilde{\mathcal{L}}_k$, by minimising the objective
\begin{equation}
\mathcal{L}_\text{hyp}(\sigma_k, \tilde{\mathcal{L}})
=
(\sigma_k - \mathrm{sg}[\tilde{\mathcal{L}}_k])^2,
\end{equation}
where $\mathrm{sg}[\cdot]$ is the stop-gradient operator, and $\tilde{\mathcal{L}}_k$ is the overall reconstruction loss under the sampled $k$-th hypothesis, given below.

\subsection{Training Objective}

Given a training sample $(I,M)$, comprising an image and the corresponding object mask, the training objective is computed in two steps.
First, the viewpoint hypothesis $k$ is sampled and the reconstruction loss is computed:
$$
\tilde{\mathcal{L}}_k =
\lambda_\text{im} \mathcal{L}_\text{im}
+ \lambda_\text{m} \mathcal{L}_\text{m}
+ \lambda_\text{f} \mathcal{L}_\text{feat}.
$$
Then, the regularisation terms and the viewpoint hypothesis loss are added, to obtain the final objective:
$$
\mathcal{L} =
p_k \tilde{\mathcal{L}}_k
+ \lambda_\text{E} \mathcal{R}_\text{Eik}
+ \lambda_\text{d} \mathcal{R}_\text{def}
+ \lambda_\text{a} \mathcal{R}_\text{art}
+ \lambda_\text{h} \mathcal{L}_\text{hyp}.
$$
Here
$
\lambda_\text{im},
\lambda_\text{m},
\lambda_\text{f},
\lambda_\text{E},
\lambda_\text{d},
\lambda_\text{a},
\lambda_\text{h}
$
are balancing weights.

The gradient of $\mathcal{L}$ with respect to the model parameters is then computed and used to update them.
Note that only one viewpoint sample $k$ is taken for each gradient evaluation.
To further improve the sampling efficiency, we sample the viewpoint based on the learned probability distribution $p_k$%
\footnote{
We sample viewpoint
$
k^*
=
\operatorname{argmax}_k p_k
$
80\% of the time for training efficiency, and uniformly at random 20\% of the time.}.

\section{Experiments}

We conduct extensive experiments on a few animal categories, including horses, giraffes, zebras, cows and birds, and compare against prior work both qualitatively and quantitatively on standard benchmarks.
We also show that our model trained on real images generalises to abstract drawings, demonstrating the power of unsupervised learning.

\subsection{Data}
For horses, we use the horse dataset from DOVE~\cite{wu2021dove} containing $10.8$k images extracted from YouTube, and supplement it with $541$ additional images from three datasets for diversity: Weizmann Horse Database~\cite{Borenstein02}, PASCAL~\cite{Everingham15} and Horse-10 Dataset~\cite{Mathis_2021_WACV}.
For giraffes, zebras and cows, we source images from Microsoft COCO Dataset~\cite{lin2014microsoft} and keep the ones with little occlusion.
Since these datasets are relatively small, we only use them to finetune the pre-trained horse model.
For birds, we combine the DOVE dataset~\cite{wu2021dove} and CUB dataset~\cite{WahCUB_200_2011} consisting of $57.9$k and $11.7$k images respectively.
We use an off-the-shelf PointRend~\cite{kirillov2019pointrend} detector to obtain segmentation masks, crop around the objects and resize them to $256 \times 256$.
We follow the original train/test splits from CUB and DOVE, and randomly split the rest, resulting in $11.5$k/$0.8$k horse, $513$/$57$ giraffe, $574$/$64$ zebra, $719$/$80$ cow and $63.9$k/$10.5$k bird images.
We additionally collect roughly $100$ horse images from the Internet to test generalisation.

\subsection{Technical Details}
The model is implemented using a total of $8$ sub-networks.
The feature field $\psi$, template SDF $s$, albedo field $f_\text{a}$ and deformation field $f_{\Delta V}$ are implemented using MLPs, which take in 3D coordinates (concatenated with image features for albedo and deformation networks) as input.
The image encoder consists of a ViT-S~\cite{Kolesnikov21vit} architecture from DINO~\cite{caron2021dino} and two convolutional encoders $f_\text{k}$ and $f_\text{o}$ that fuse the patch features into global feature vectors.
We use self-supervised pre-trained DINO-ViT with frozen weights and only train the convolutional encoders.
The lighting network $f_\text{l}$ is a simple MLP that maps global output feature to a $4$-channel vector, and the articulation network $f_\text{b}$ is a transformer architecture with $4$ blocks.
We use a separate encoder for the viewpoint network $f_\text{vp}$ identical to $f_\text{k}$, as we find empirically that sharing the encoders tends to make the viewpoint learning unstable.
Apart from DINO encoder, all components are trained end-to-end from scratch for $100$ and $10$ epochs on horses and birds respectively, with the instance deformation and articulation disabled for the first $30$ and $2$ epochs.
The DINO features $\Phi'_\text{k}$ used to compute $\mathcal{L}_\text{feat}$ are pre-computed with patch size $8$ and stride $4$ and masked, following~\cite{amir2021deep}, and reduced from a channel size of $384$ to $16$ using PCA\@.
All technical details and hyper-parameters are included in \ifarxiv \cref{sec:supmat_tech_details}. \else the supplementary material. \fi

\begin{figure}[t]
    \centering
    \includegraphics[trim={0 0 20px 0}, clip, width=\linewidth]{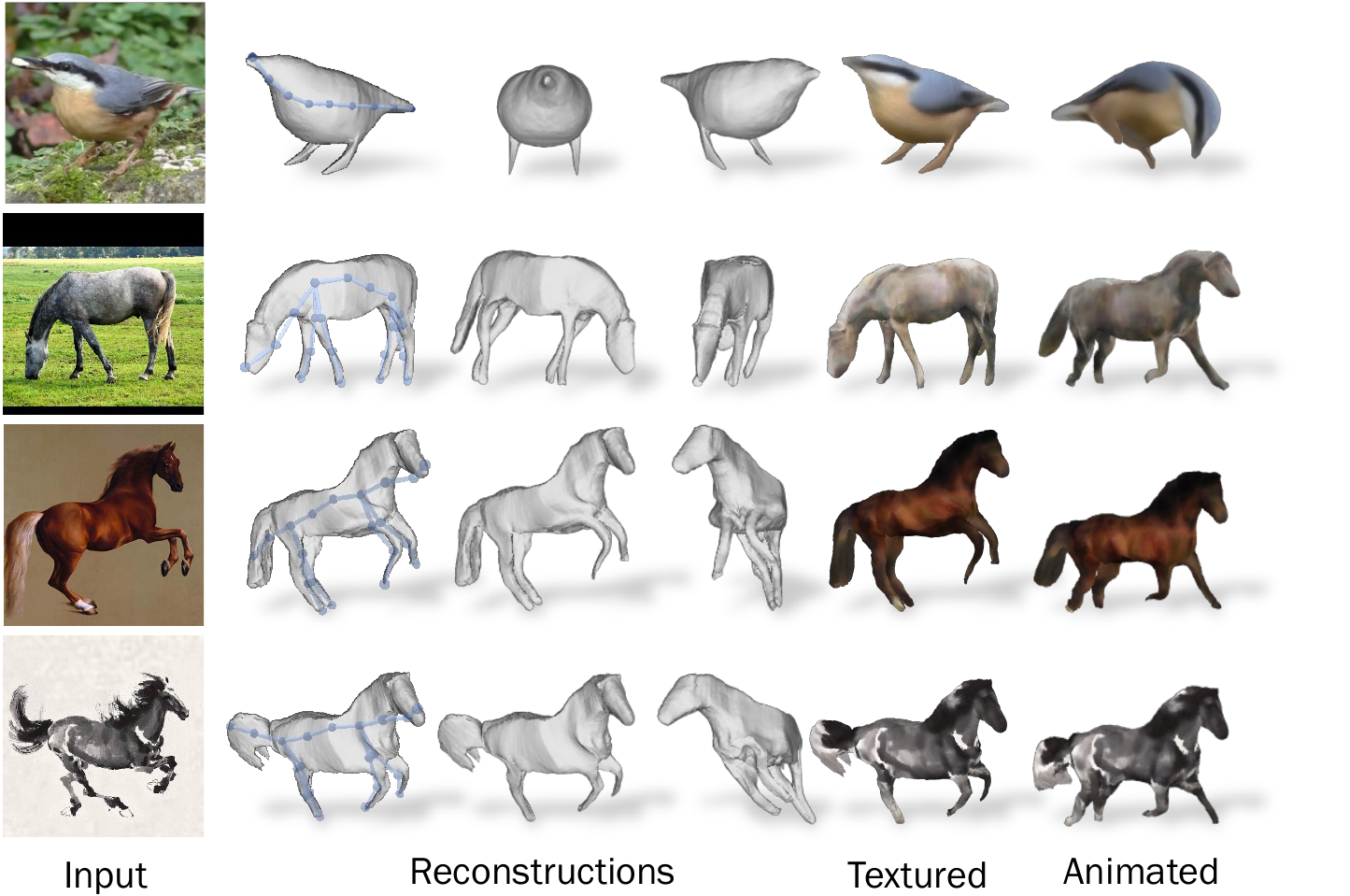}
    \caption{\textbf{Single Image Reconstruction.}
    We show the reconstructed mesh from the input view and two additional views together with predicted texture and animated version of the shape obtained by articulating the estimated skeleton.
    }%
    \label{fig:recon}
\end{figure}
\begin{figure}[t]
    \centering
    \includegraphics[trim={0 0 20px 0}, clip, width=\linewidth]{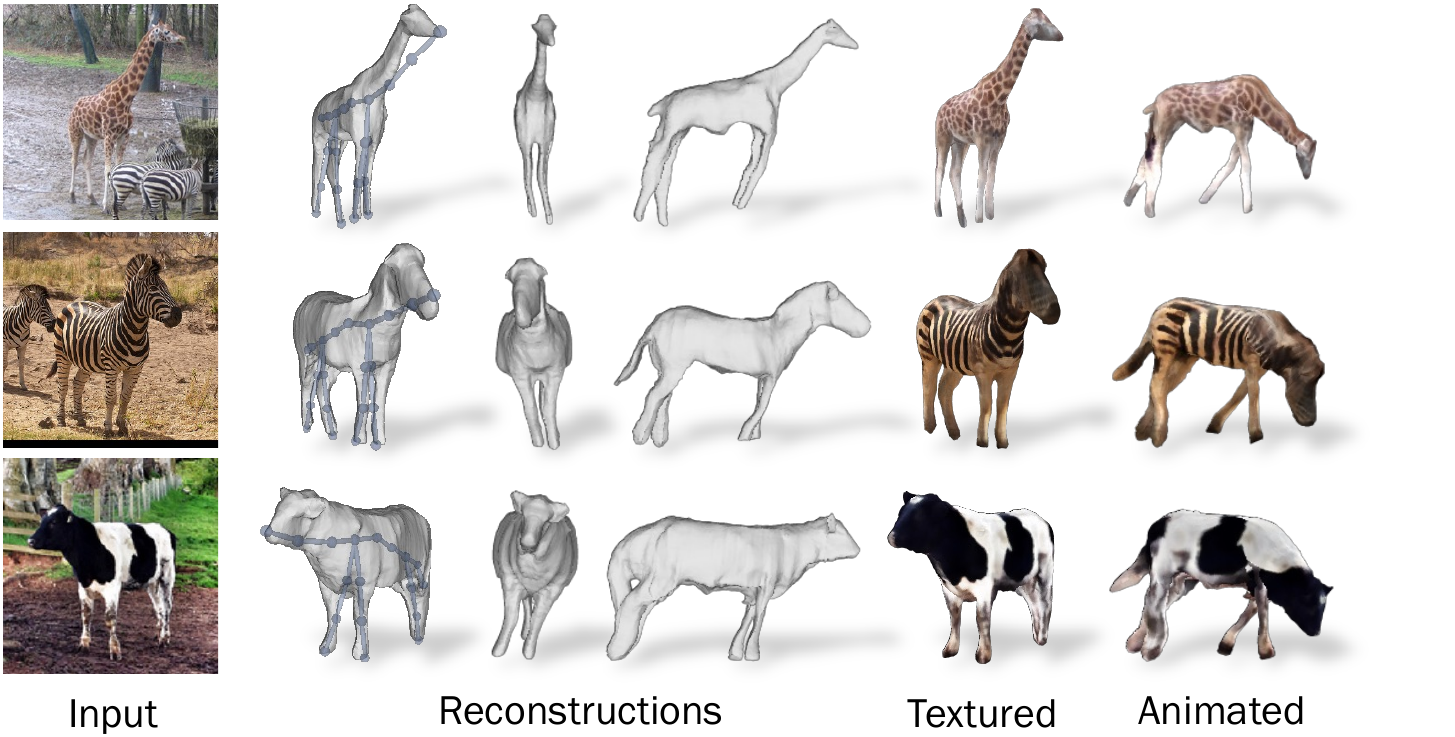}
    \caption{\textbf{Reconstructions of Giraffes, Zebras and Cows.}
    We finetune the horse model on other animal categories, and show that the method generalises well to other animals.
    }%
    \label{fig:other_animals}
\end{figure}

\subsection{Qualitative Results}
\cref{fig:recon} shows a few reconstructions of horses and birds produced by our model.
Given a single 2D image at test time, our model predicts a textured 3D mesh of the object, capturing its fine-grained geometric details, such as the legs and tails of the horses.
Our model predicts the articulated pose of the objects, allowing us to easily transfer the pose of one instance from another and animate it in 3D.
Although our model is trained on real images only, it demonstrates excellent generalisation to paintings and abstract drawings.

\cref{fig:other_animals} also shows that by finetuning the horse model on other animals without any additional modifications (except disabling articulation for the initial $5$k iterations), it generalises to various animal categories with highly different shapes and appearances, such as giraffes, zebras and cows.

Note that for the examples of abstract horse drawings (last rows in \cref{fig:teaser,fig:recon}), since the texture is out of the training distribution, we finetune (only) the albedo network for $100$ iterations, which takes less than $10$ seconds.
This is also done for giraffes, zebras and cows in \cref{fig:other_animals}, as the training datasets are too small to learn complex appearance.
Additional qualitative results are presented in \ifarxiv \cref{sec:supmat_results}. \else the sup.~mat. \fi

\begin{figure}[t]
    \centering
    \includegraphics[trim={0 0 30px 0}, clip, width=\linewidth]{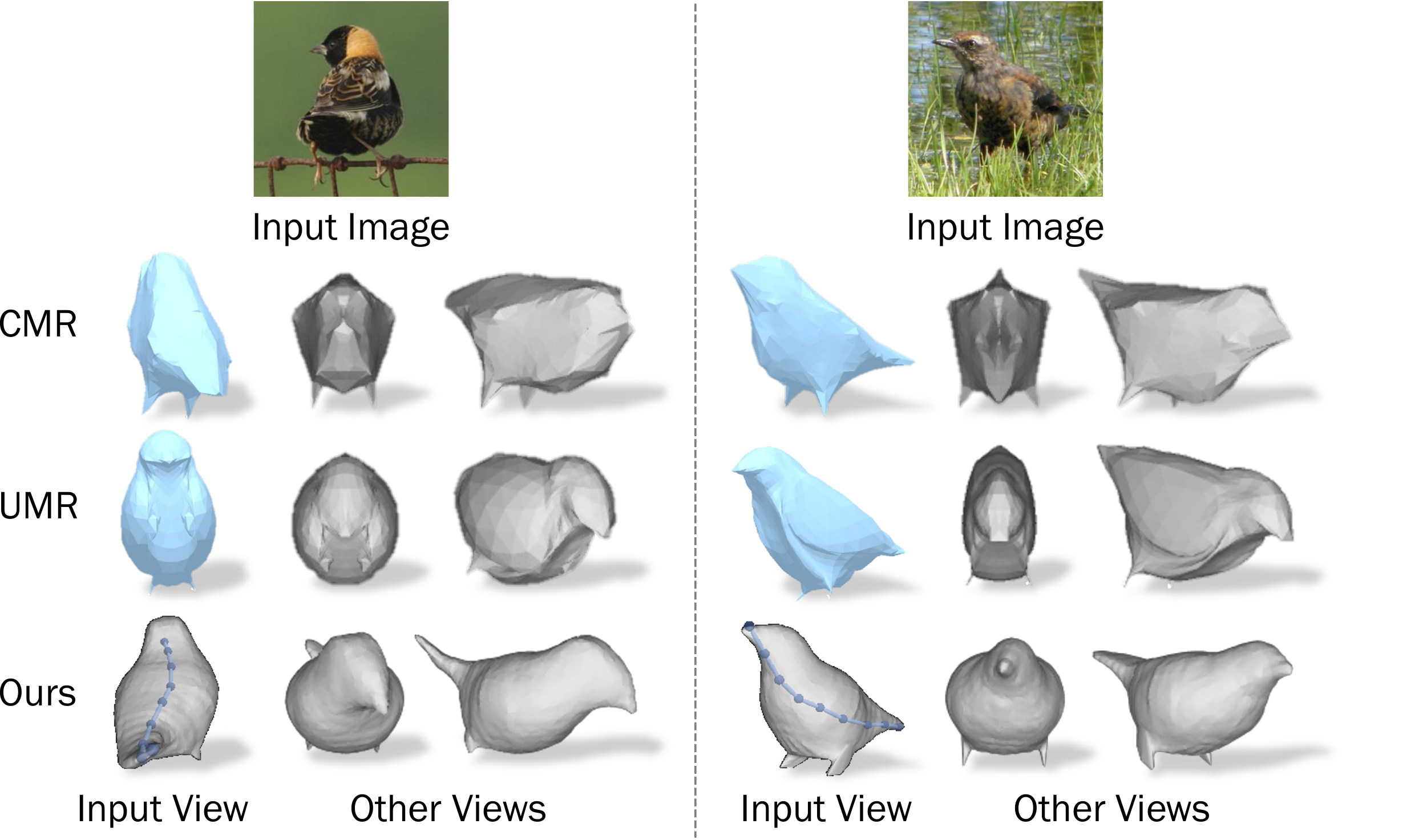}
    \caption{\textbf{Comparison on Birds.} Both CMR~\cite{kanazawa18cmr} and UMR~\cite{Li2020umr} often predict inaccurate poses, such as the bird on the left.}%
    \label{fig:compare_bird}
\end{figure}
\begin{figure}[t]
    \centering
    \includegraphics[trim={0 0 20px 0}, clip, width=\linewidth]{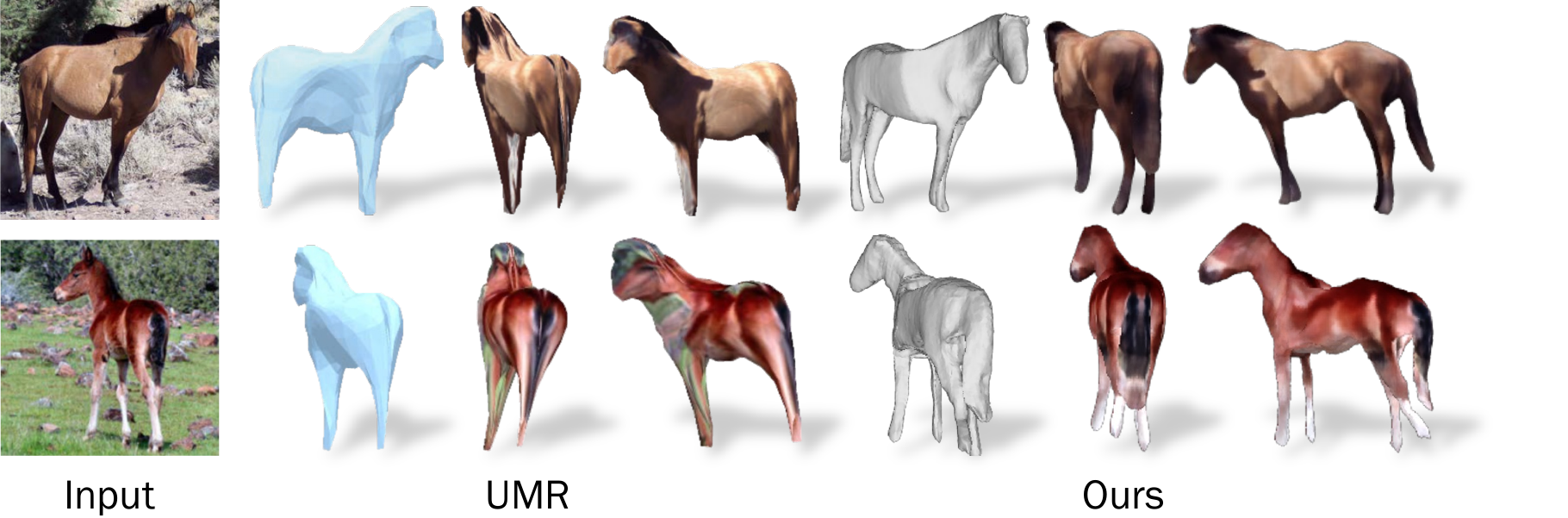}
    \caption{\textbf{Comparison with UMR~\cite{Li2020umr} on Horses.}
    Our model produces much higher quality reconstructions with four legs.
    UMR is only able to predict symmetrical shapes with two legs.}%
    \label{fig:compare}
\end{figure}

\subsection{Comparison with Prior Work}
\label{sec:compare}
We compare with previous weakly-supervised methods for 3D reconstruction of deformable objects.
The most relevant prior work is UMR~\cite{Li2020umr} and DOVE~\cite{wu2021dove}, as they also only require object masks and weak correspondences from either part segmentations (SCOPS)~\cite{hung19scops} or video training~\cite{wu2021dove}.
Our method leverages \emph{self-supervised} DINO-ViT features, which makes our method the least supervised in this area.
We also compare against the well-established baseline of CMR~\cite{kanazawa18cmr}, which requires stronger supervision in the form of 2D keypoint annotations and pre-computed cameras, as well as U-CMR~\cite{Goel20ucmr}, which uses a category-specific template shape.
To evaluate other methods, we use their code and pre-trained models in public repositories (CMR, UMR, U-CMR) or provided by the authors (DOVE).

\paragraph{Qualitative Comparisons.}
\cref{fig:compare,fig:compare_bird} compare our reconstructions against the previous methods on bird and horse images.
Since UMR did not release their horse model, we test our model on the examples presented in their paper for side-by-side comparison.
Our method predicts more accurate poses and higher-quality articulated shapes (as opposed to only symmetric shapes with CMR and UMR).
On horses, our method reconstructs all four legs with details while UMR predicts only two.
Since UMR copies texture from the input image, for a fair comparison, we finetune (only) the albedo network for $100$ iterations at test time.

\begin{table} [t!]
    \footnotesize
    \newcommand{\xpm}[1]{{\tiny$\pm#1$}}
	\centering
    \caption{\textbf{Evaluation on Toy Bird Scans~\cite{wu2021dove}.}
    Baseline results reported by~\cite{wu2021dove}.
    Annotations: \stemp template shape, \sview viewpoint, \skey 2D keypoint, \smask mask, \sflow optical flow, \svid video.}
	\begin{tabular}{lcc}
		\toprule
		 & Supervision & Chamfer Distance (cm) $\downarrow$  \\
		\midrule
		 CMR~\cite{kanazawa18cmr} & \supcmr   & 1.35 \xpm{0.81} \\
		  U-CMR~\cite{Goel20ucmr} & \supucmr & 1.82 \xpm{0.93} \\
		 UMR~\cite{Li2020umr} & \supumr+ SCOPS   & 1.24 \xpm{0.75}  \\ %
		 DOVE~\cite{wu2021dove}  & \supdove & 1.51 \xpm{0.89}  \\ 
		 \midrule
		 Ours  & \supours+ DINO & \textbf{0.79} \xpm{0.50}  \\ 
		\bottomrule
	\end{tabular}%
	\label{tab:toy_bird}
\end{table}

\begin{table} [t!]
  \footnotesize
  \newcommand{\xpm}[1]{{\tiny$\pm#1$}}
\centering
\setlength{\tabcolsep}{7pt}
  \caption{\textbf{Keypoint Transfer on PASCAL~\cite{Everingham15} Horse and CUB~\cite{WahCUB_200_2011} Bird Datasets.}
  Our model produces superior results on both categories with significantly less supervision.
  $^\dagger$: numbers taken from \cite{Li2020umr}.
  `Birds w/o aq.': excluding 50 aquatic bird classes (\eg mallards) with heavy occlusion and large shape variation.
  }
\begin{tabular}{lcccc}
  \toprule
  & & \multicolumn{3}{c}{PCK@0.1  $\uparrow$} \\
	\cmidrule(lr){3-5}
  Method & Supervision & Horses & Birds & Birds w/o aq. \\
  \midrule
      CMR~\cite{kanazawa18cmr}$^\dagger$ & \supcmr & -- & 0.473 & -- \\
      CMR~\cite{kanazawa18cmr} & \supcmr & -- & 0.546 & 0.591 \\
	   CMR~\cite{kanazawa18cmr} & \smask & -- & 0.255 & 0.277 \\
      U-CMR~\cite{Goel20ucmr} & \supucmr & -- & 0.359 & 0.412\\
      UMR~\cite{Li2020umr} & \smask+ SCOPS & 0.284 & 0.512 & 0.555 \\
      A-CSM~\cite{kulkarni20articulation-aware} & \supacsm & 0.329 & 0.426 & -- \\
      DOVE~\cite{wu2021dove} & \supdove & -- & 0.447 & 0.510 \\
   \midrule
   Ours & \supours+ DINO & \textbf{0.429} & \textbf{0.555} & \textbf{0.635} \\
  \bottomrule
\end{tabular}%
\label{table:cub}
\end{table}

\paragraph{Quantitative Comparisons.}
We evaluate shape reconstruction on the 3D Toy Bird Dataset from DOVE~\cite{wu2021dove}, which contains 3D scans of $23$ realistic toy bird models paired with $345$ photographs of them taken in natural real-world environments.
Following~\cite{wu2021dove}, we align the predicted mesh with the ground-truth scan using ICP and compute the bi-directional Chamfer Distance between two sets of sample points from the aligned meshes.
\cref{tab:toy_bird} summarises the results compared to other methods.
Our model produces significantly more accurate reconstructions, resulting in a much lower error.

    We also evaluate Keypoint Transfer on the CUB~\cite{WahCUB_200_2011} and PASCAL VOC~\cite{Everingham15} benchmarks, a common evaluation metric~\cite{kanazawa18cmr, Li2020umr, kulkarni20articulation-aware}. 
We follow the protocols in~\cite{Li2020umr} and~\cite{kulkarni20articulation-aware} for birds and horses, respectively, and sample 20k source and target image pairs.
Given a source image, we project all the visible vertices of the predicted mesh onto the image using the predicted viewpoint
and assign each annotated 2D keypoint to its nearest vertex.
We then render these vertices using the mesh and viewpoint predicted from a different target image.
We measure the error between the projected vertices (corresponding to the transferred keypoints) and the annotated target keypoints using the Percentage of Correct Keypoints (PCK) metric.
As shown in \cref{table:cub},
our model significantly outperforms previous methods relying on stronger geometric supervision, including A-CSM~\cite{kulkarni20articulation-aware} which requires a 3D shape template to begin with.

\begin{figure}[t]
    \centering
    \includegraphics[trim={0 0 30px 0}, clip, width=\linewidth]{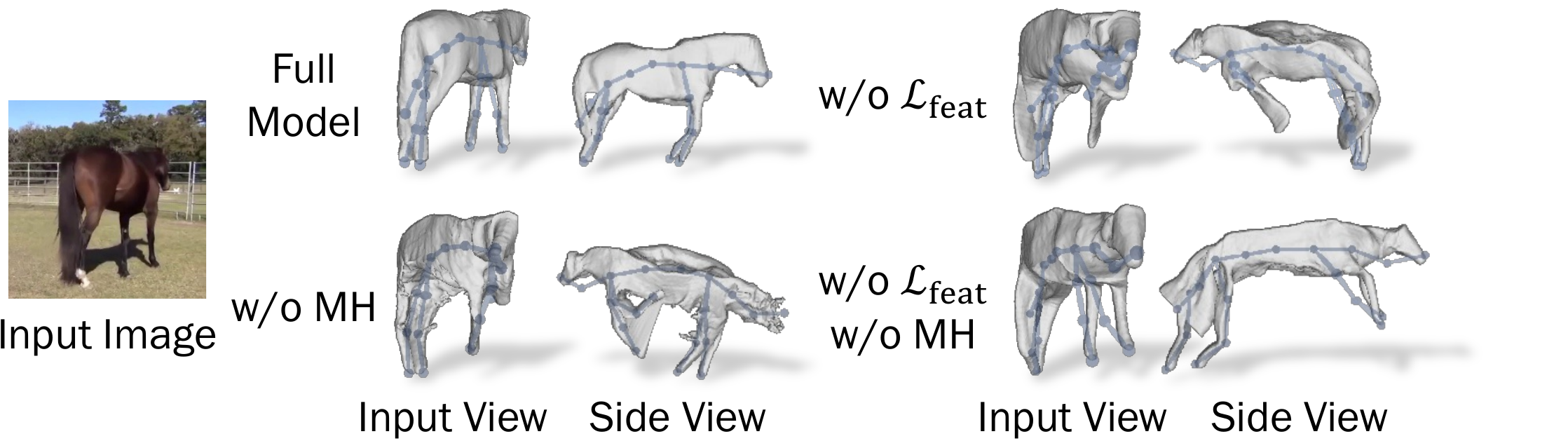}
    \caption{\textbf{Ablation Studies.}
    We train the model without feature loss $\mathcal{L}_\text{feat}$, without multi-hypothesis (MH) viewpoint prediction and without both.
    Without either component, the viewpoint predictions collapse to frontal views, resulting in unnatural shapes.}%
    \label{fig:ablation}
\end{figure}

\subsection{Ablation Studies and Limitations}
We present ablation studies on the two key components of the model in \cref{fig:ablation}, self-supervised feature loss and multi-hypothesis viewpoint prediction.
Without either component, the learned viewpoints collapse to only frontal views, resulting in unnaturally stretched reconstructions.
Additional results are provided in \ifarxiv \cref{fig:supmat_abl_view}. \else the sup.~mat. \fi

Although our model reconstructs highly detailed 3D shapes from only a single image and generalises to in-the-wild images, the predicted texture may not preserve sufficient details with a single forward pass and may require additional test-time finetuning.
This can potentially be addressed by training on a larger dataset and incorporating recent advances in image generation~\cite{rombach2021highresolution, saharia2022photorealistic}.
Another limitation is the requirement of a pre-defined topology for articulation, which may vary between different animal species.
Discovering the articulation structure automatically from raw in-the-wild images will be of great interest for future work.
Failure cases are discussed in \ifarxiv \cref{sec:supmat_failure}. \else the sup.~mat. \fi

\section{Conclusions}%
\label{s:conclusions}

We have introduced a new model that can learn a 3D model of an articulated object category from single-view images taken in the wild.
This model can, at test time, reconstruct the shape, articulation, albedo, and lighting of the object from a single image, and generalises to abstract drawings.
Our approach demonstrates the power of combining several recent improvements in self-supervised representation learning together with a new viewpoint sampling scheme.
We have shown significantly superior results compared to prior works, even when they use more supervision.

\paragraph{Data Ethics.}

We use the DOVE, Weizmann Horse, PASCAL, Horse-10, and MS-COCO datasets in a manner compatible with their terms.
Some of these images may accidentally contain faces or other personal information, but we do not make use of these images or image regions.
For further details on ethics, data protection, and copyright please see {\small\url{https://www.robots.ox.ac.uk/~vedaldi/research/union/ethics.html}}.

\paragraph{Acknowledgements.}
Shangzhe Wu is supported by a research gift from Meta Research.
Andrea Vedaldi and Christian Rupprecht are supported by ERC-CoG UNION 101001212.
Tomas Jakab and Christian Rupprecht are (also) supported by VisualAI EP/T028572/1.

{\small\bibliographystyle{ieee_fullname}\bibliography{ref}}

\newpage

\section{Additional Results}
\label{sec:supmat_results}

\subsection{Additional Comparisons with Prior Works}
\label{sec:supmat_compare}

\paragraph{Qualitative Comparisons}
\cref{fig:supmat_compare_horse} and \cref{fig:supmat_compare_bird} show qualitative comparisons with prior works on both horses and birds.
Our method predicts 3D shapes with finer details and more accurate poses, compared to prior works.
We also plot the distribution of predicted viewpoints demonstrating that other methods with a comparable level of supervision collapse to only a limited range of viewpoints, \eg predicting only frontal poses.

We also compare against another recent method, LASSIE~\cite{yao2022lassie}, which also leverages DINO-ViT~\cite{caron2021dino} image features but only \emph{optimises} over a small set of images ($\sim30$).
As illustrated in \cref{fig:supmat_compare_lassie}, LASSIE~\cite{yao2022lassie} starts from a heavily hand-crafted part-based initial shape, whereas our method starts with a generic ellipsoid with only a simple description of the bone topology.
Yet after training, our model produces more detailed 3D shapes from a new test image \emph{unseen} at training, compared to the reconstructions obtained by LASSIE which are directly \emph{optimised} on these images.

\paragraph{Visualisations of Toy Bird Reconstructions.}
Supplementary to \ifarxiv \cref{tab:toy_bird}, \else Tab.~2 in the main paper, \fi we show a qualitative comparison of the predicted shapes and the scanned ground-truth shapes on the Toy Bird Scan dataset in~\cref{fig:supmat_compare_toy_bird}.

\subsection{Ablation Studies}
We further provide quantitative ablation studies on the Toy Bird Scans benchmark in \cref{tab:supmat_ablation}, validating the effectiveness of individual components of the model.
In addition to \ifarxiv \cref{fig:ablation}, \else Fig.~5 in the main paper, \fi we also examine the effects of both the feature rendering loss $\mathcal{L}_\text{feat}$ and the multi-hypothesis viewpoint prediction in \cref{fig:supmat_abl_view}, demonstrating that both of the components are essential to prevent the collapse of viewpoint prediction.

\subsection{Texture Finetuning}
\cref{fig:supmat_texture_finetune} shows how a quick test-time finetuning (100 iterations) of the predicted texture improves their quality.
This is especially effective for images that are far from the training set distribution.
Note that the textures of the real horses in the main paper are predictions from a single forward pass.

\begin{figure}[t]
    \centering
    \includegraphics[trim={0 0 20px 0}, clip, width=\linewidth]{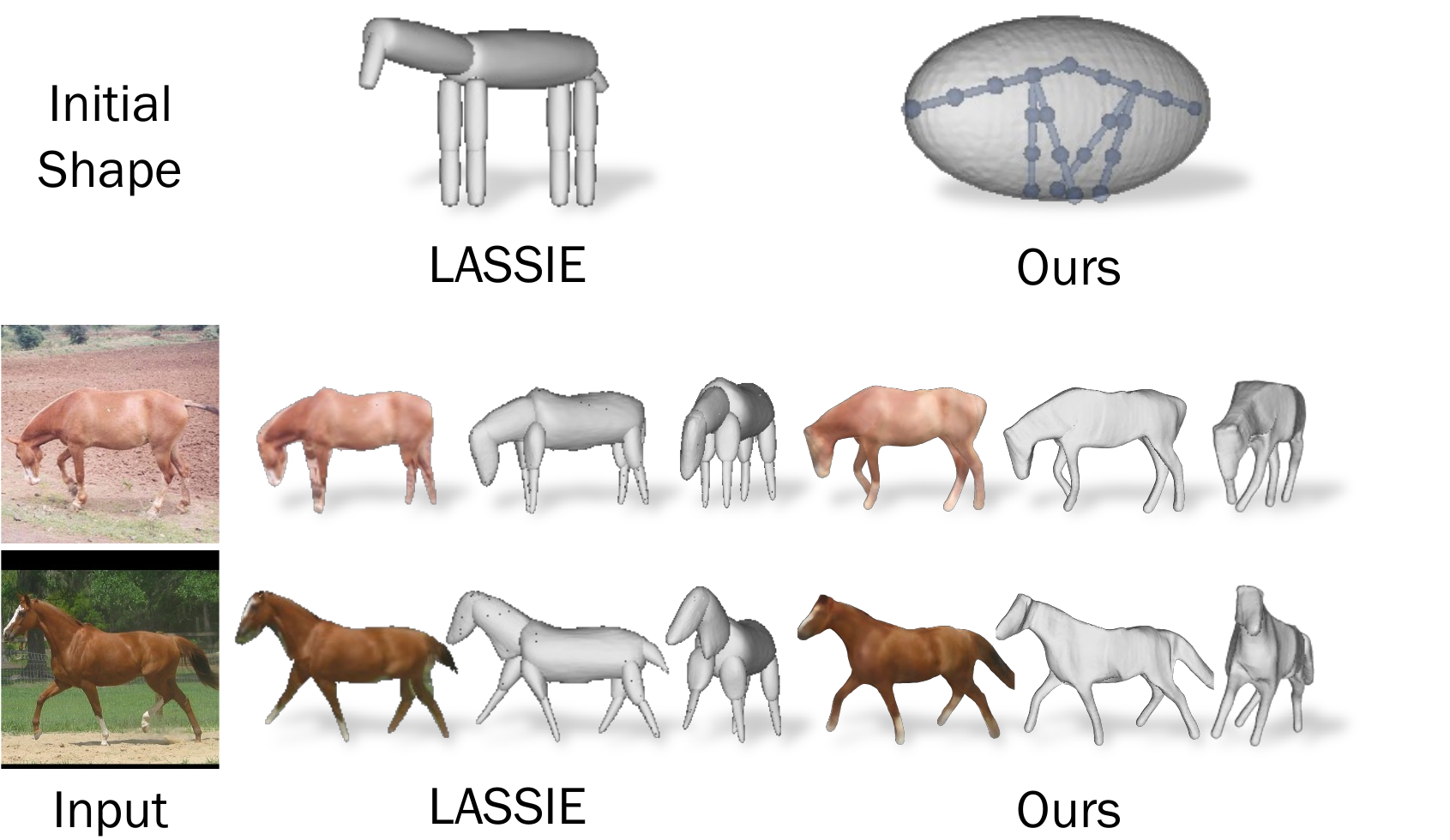}
    \caption{\textbf{Comparison with LASSIE~\cite{yao2022lassie} on Horses.}
    LASSIE starts from a heavily hand-crafted part-based initial shape, whereas our method simply starts with a generic ellipsoid with a simple heuristic description of the bone topology.
    After training, our method produces more plausible 3D shapes from an \emph{unseen} test image, compared to the reconstructions obtained from LASSIE which are directly \emph{optimised} on these images.
    Note that LASSIE represents the 3D shape with a set of disjoint primitive parts, resulting in unnatural junctions.
    }
    \label{fig:supmat_compare_lassie}
\end{figure}

\subsection{Additional Qualitative Results}
Additional results of single image reconstruction, animation and relighting can be found in \cref{fig:supmat_recon_horse_real} and the supplementary video.
More generalisation results on abstract horse drawings, sculptures and toys are presented in \cref{fig:supmat_recon_horse_abstract}, showing that the model has learned to estimate shape, pose and articulation sufficiently robustly to generalise beyond the training distribution.

We also show more reconstruction results of giraffes, zebras and cows in \cref{fig:supmat_recon_other_animals}.
Our method is able to produce accurate shape reconstructions from a single image across large variations of animal shapes.

\subsection{Failure Cases}
\label{sec:supmat_failure}
Our texture prediction might not generalise well enough beyond the distribution of textures observed during training.
This is particularly apparent when the trained model is applied on paintings and abstract drawings of horses, neither of which are part of the training set.
We demonstrate that a quick finetuning step of the albedo network ($100$ iterations which takes less than $10$ seconds) can remedy this shortcoming.
\cref{fig:supmat_texture_finetune} illustrates the difference between the single-pass predicted textures and the finetuned version.

\begin{figure}[t]
    \centering
    \includegraphics[trim={0 0 40px 0}, clip, width=\linewidth]{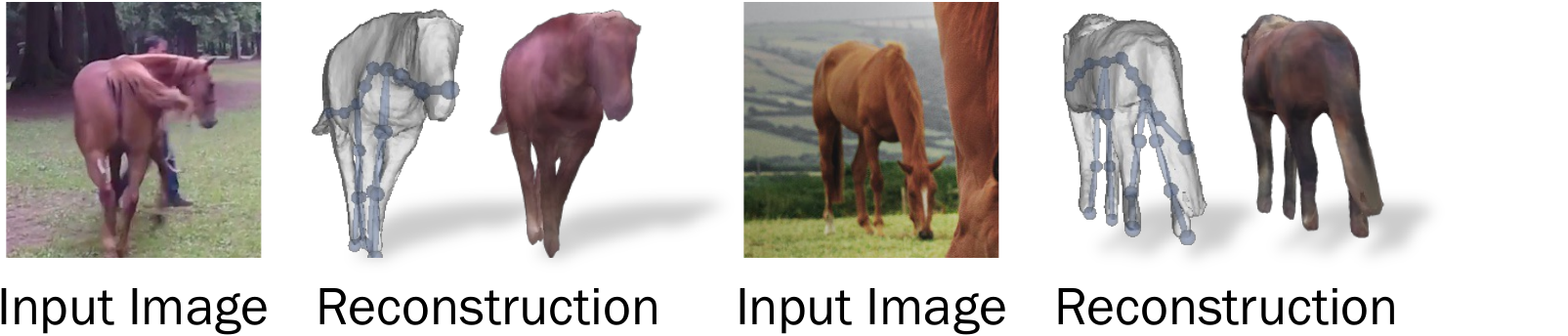}
    \caption{\textbf{Incorrect Viewpoint Predictions.}
    The viewpoint prediction can be less reliable in the case of more extreme input views.
    }
    \label{fig:supmat_failure_viewpoint}
\end{figure}

\begin{figure*}[t]
    \centering
    \includegraphics[width=\linewidth]{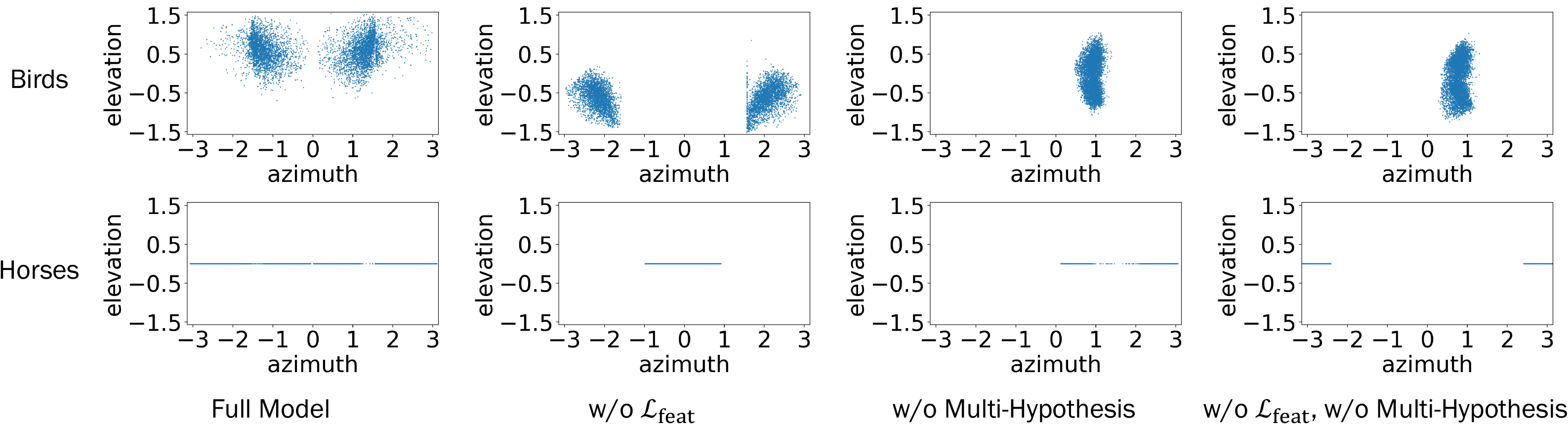}
    \caption{\textbf{Visualisations of the Viewpoint Prediction Distributions of the Ablated Models.}
    We demonstrate that both feature reconstruction loss $\mathcal{L}_\text{feat}$ and multi-hypothesis viewpoint prediction are needed to successfully recover a full range of viewpoints.
    The viewpoint prediction collapses to a limited range as demonstrated by its azimuth without these two components.
    Note that for Horses, we only predict the azimuth of the viewpoint, as most of the horse images were taken with little elevation.
    }
    \label{fig:supmat_abl_view}
\end{figure*}

\begin{figure*}[t]
    \centering
    \includegraphics[trim={0 0 20px 0}, clip, width=\linewidth]{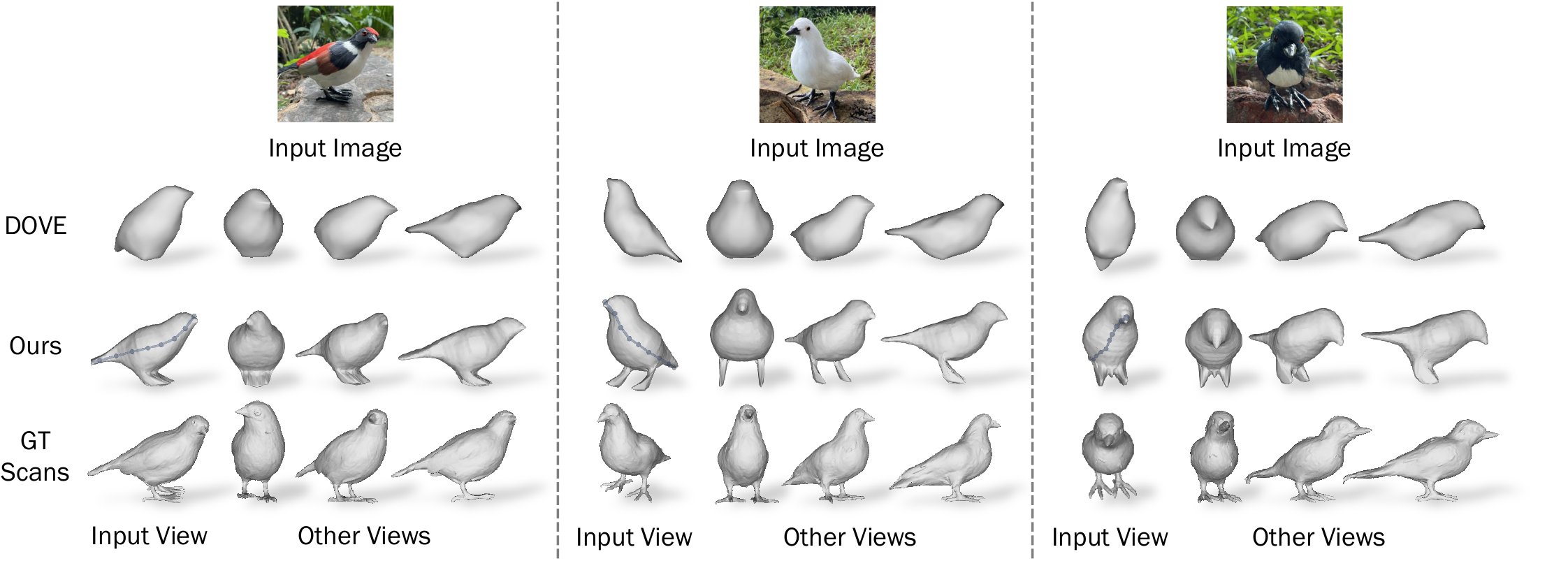}
    \caption{\textbf{Visual Comparison on Toy Bird Scans Evaluations.}
    We compare the reconstructed shapes with scanned ground-truth shapes from Toy Bird Scans dataset.
    We show the reconstructed mesh from the input view and three additional views.
    Our model is able to predict finer shape details including the bird's legs as opposed to the prior work of DOVE~\cite{wu2021dove}.
    }
    \label{fig:supmat_compare_toy_bird}
\end{figure*}

\begin{figure*}[t]
    \centering
    \includegraphics[trim={0 0 20px 0}, clip, width=\linewidth]{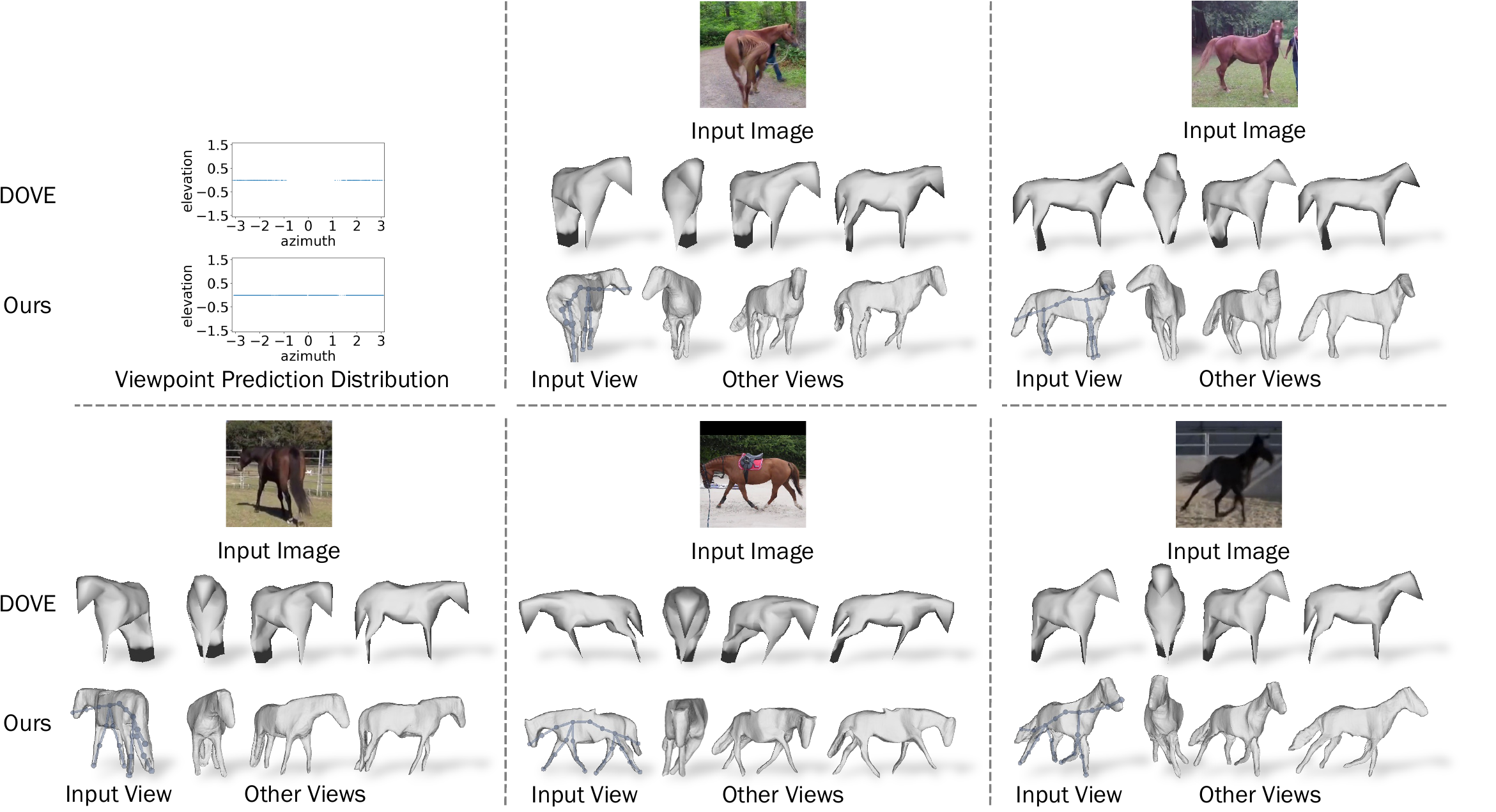}
    \caption{\textbf{Comparison with DOVE~\cite{wu2021dove} on Horses.}
    We visualise the distribution of predicted viewpoints on the test set together with additional qualitative results.
    Our method is able to recover the full range viewpoint azimuth, while DOVE covers only a portion of possible azimuths.
    This is further illustrated by the qualitative results, where DOVE often fails to predict the correct viewpoint as opposed to our method.
    Moreover, our predicted shape is far more detailed.
    Note that for horses, we only predict azimuth rotations, as most of the horse images were taken with little elevation.
    }
    \label{fig:supmat_compare_horse}
\end{figure*}

\begin{figure*}[t]
    \centering
    \includegraphics[trim={0 0 20px 0}, clip, width=\linewidth]{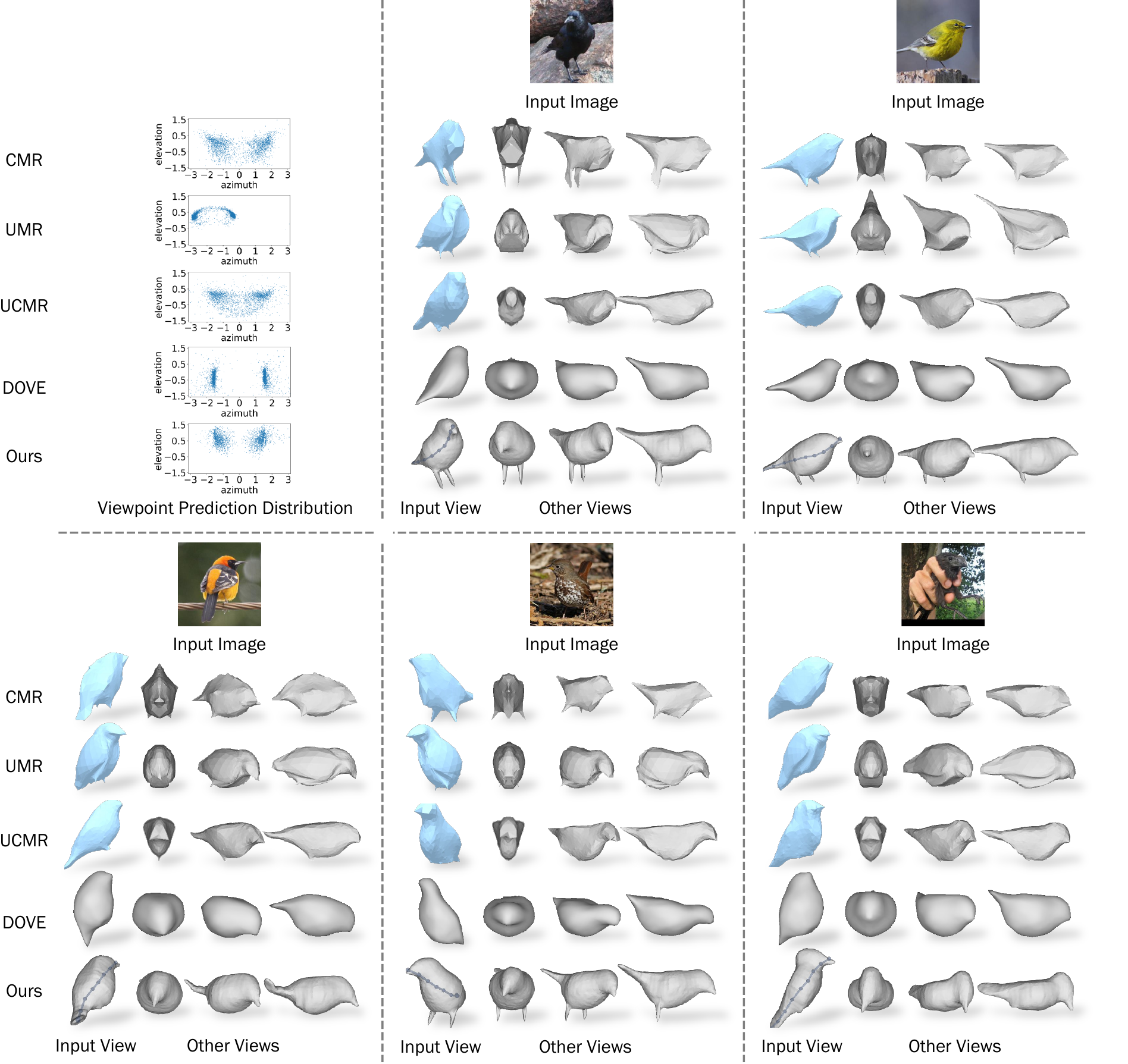}
    \caption{\textbf{Comparison with Previous Methods on Horses.}
    As in~\cref{fig:supmat_compare_horse} we visualise the distribution of predicted viewpoints on the test set together with additional qualitative results.
    The plot of viewpoint prediction distribution on CUB test set shows that our method is able to recover a wide range of viewpoints while UMR, which uses a similar level of supervision, is able to predict only frontal poses.
    We also present additional qualitative results on CUB test set demonstrating that our method recovers shapes with greater details than previous works while using significantly less supervision.
    }
    \label{fig:supmat_compare_bird}
\end{figure*}

\begin{figure*}[t]
    \centering
    \includegraphics[trim={0 0 20px 0}, clip, width=\linewidth]{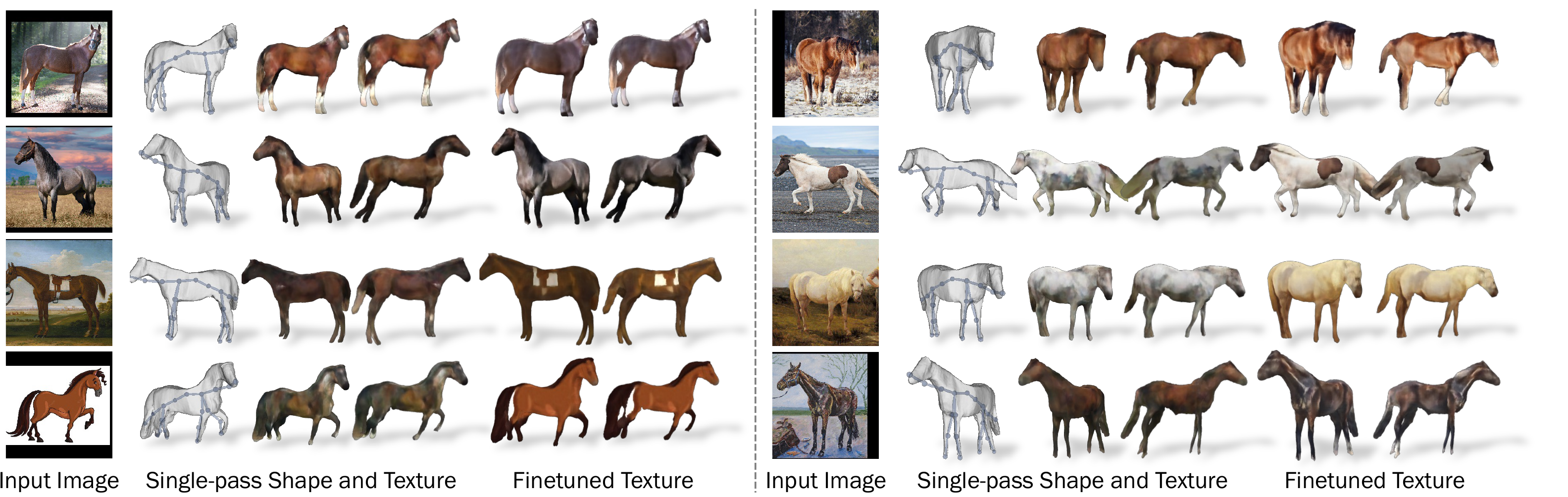}
    \caption{\textbf{Texture Finetuning at Test Time.}
    We show a shape and texture prediction from an input view and one additional view together with a finetuned version of the texture.
    We demonstrate that a simple finetuning of the texture on the input image can produce high-quality textures for images that are too far from the training set distribution.
    }
    \label{fig:supmat_texture_finetune}
\end{figure*}

The viewpoint prediction can fail in the case of more extreme and ambiguous views as shown~\cref{fig:supmat_failure_viewpoint}.
This is often caused by DINO-ViT features that are less distinctive for these particular views.

\begin{figure*}[t]
    \centering
    \includegraphics[trim={0 0 20px 0}, clip, width=\linewidth]{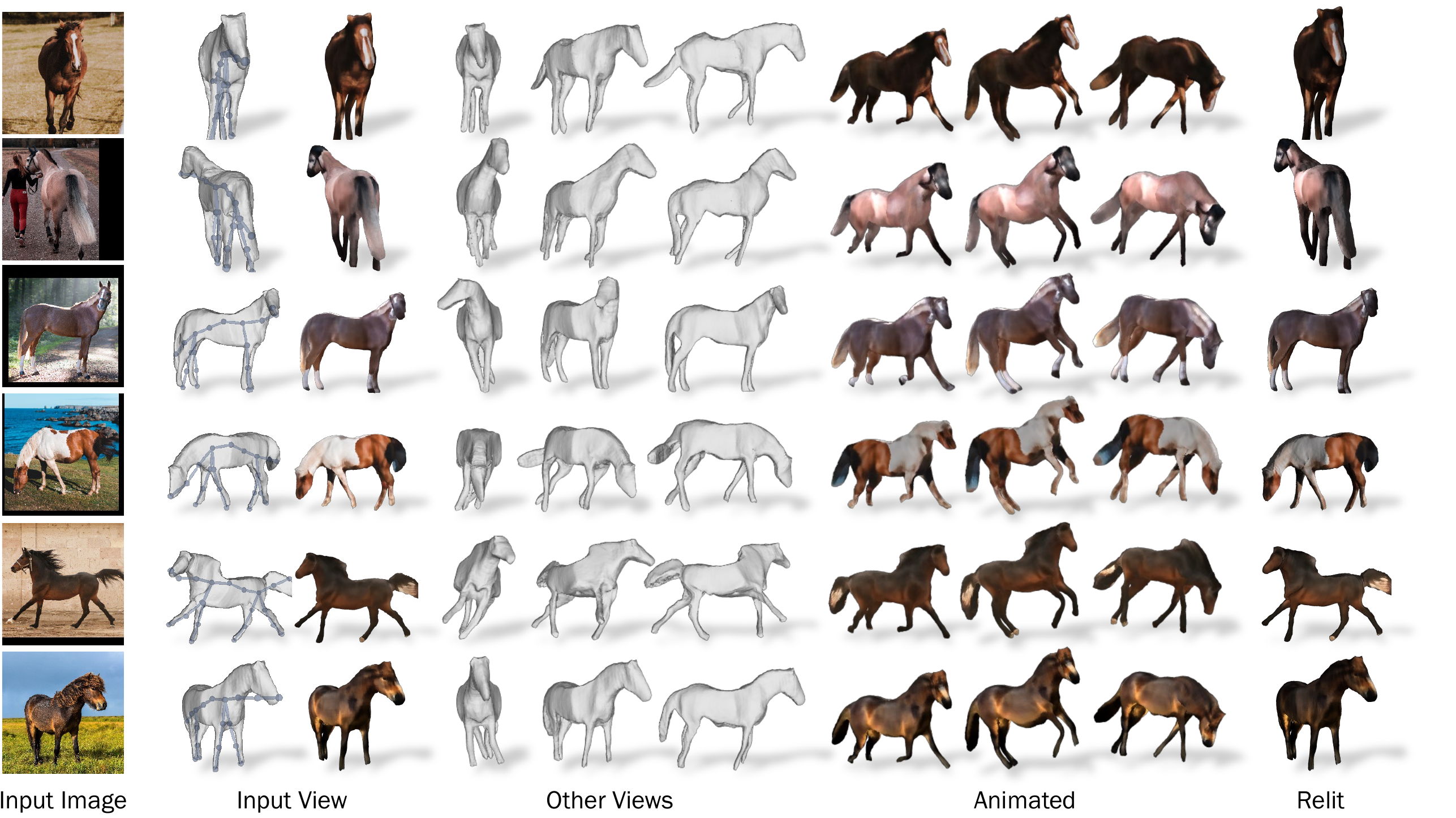}
    \caption{\textbf{Reconstruction of Real Horse Images.}
    We show the predicted mesh from the input view and three additional views.
    We also demonstrate that our shape can be animated by articulating the estimated skeleton.
    Finally, as our method decomposes albedo and lightning, our predictions can be easily relit.
    }
    \label{fig:supmat_recon_horse_real}
\end{figure*}

\begin{figure*}[t]
    \centering
    \includegraphics[trim={0 0 20px 0}, clip, width=\linewidth]{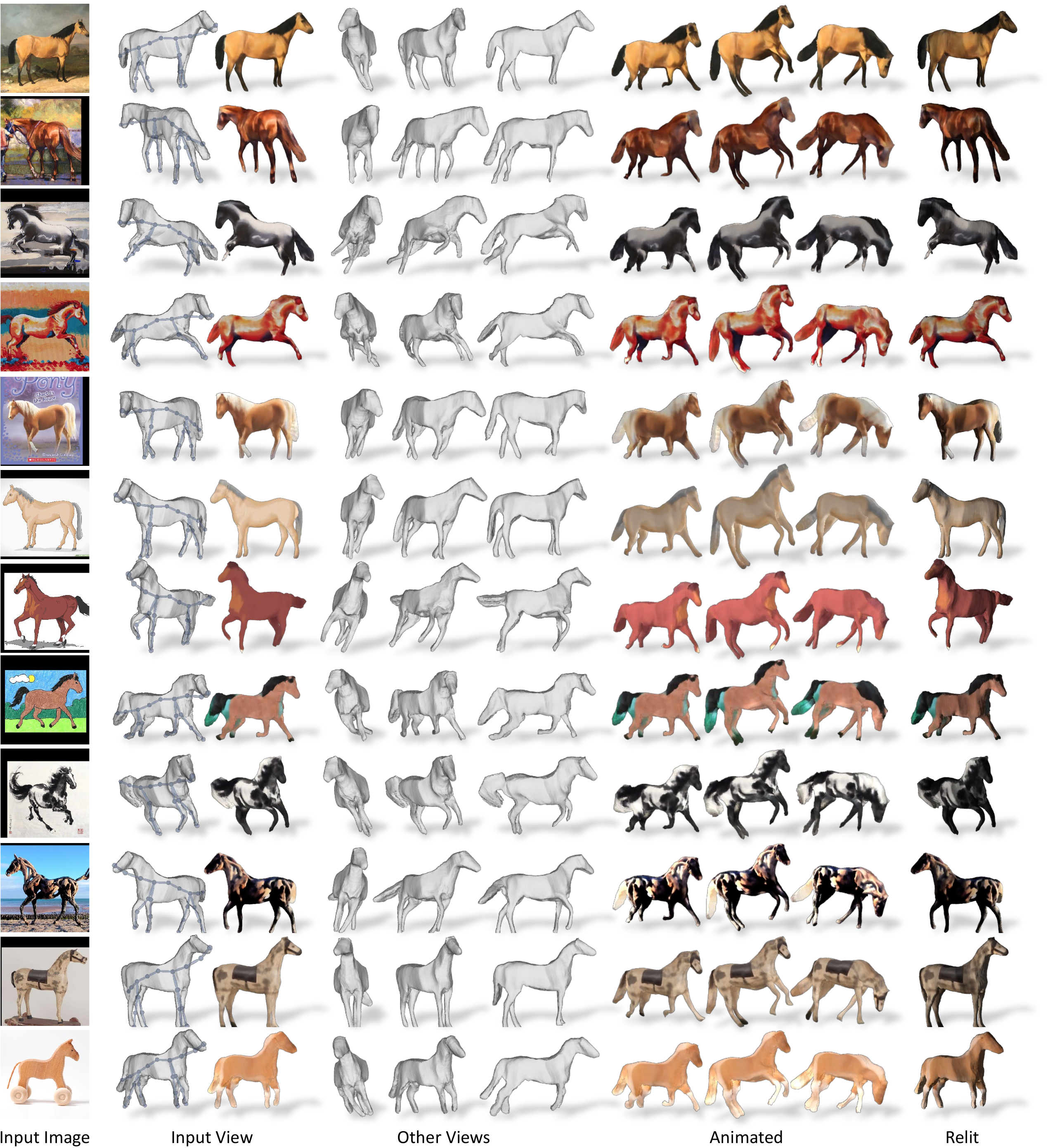}
    \caption{\textbf{Reconstruction of Abstract Horse Drawings and Artefacts.}
    As in~\cref{fig:supmat_recon_horse_real}, here we show the predicted meshes from the input view and three additional views together with the animated and relit versions.
    The results demonstrate excellent generalisation of our method on images far from the distribution of the training set which consists only of real horse images.
    }
    \label{fig:supmat_recon_horse_abstract}
\end{figure*}

\begin{figure*}[t]
    \centering
    \includegraphics[trim={0 0 20px 0}, clip, width=\linewidth]{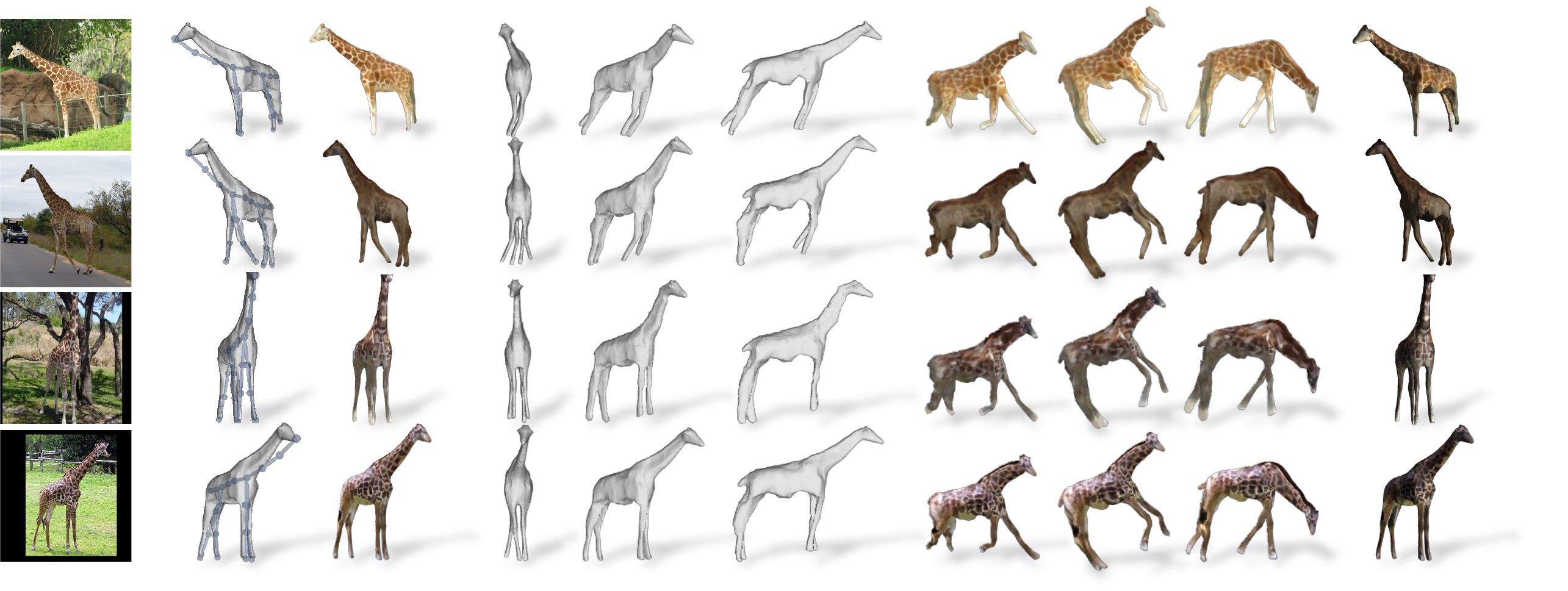}
    \includegraphics[trim={0 0 20px 0}, clip, width=\linewidth]{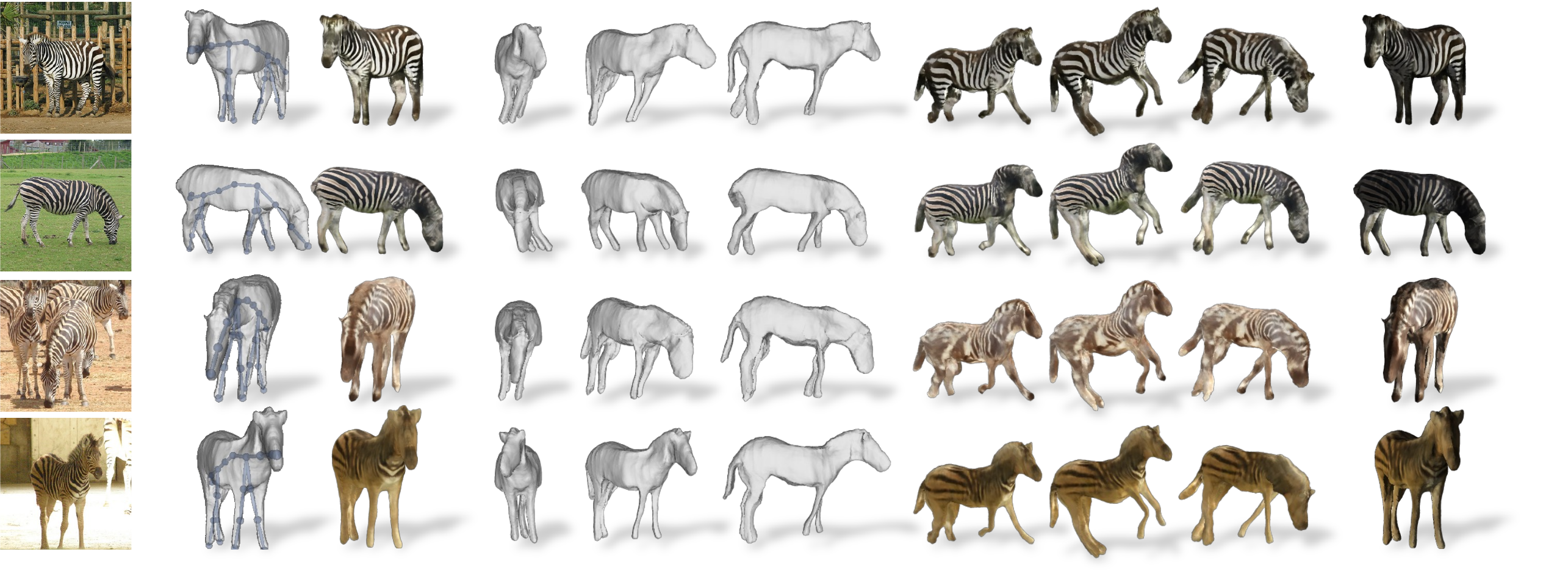}
    \includegraphics[trim={0 0 20px 0}, clip, width=\linewidth]{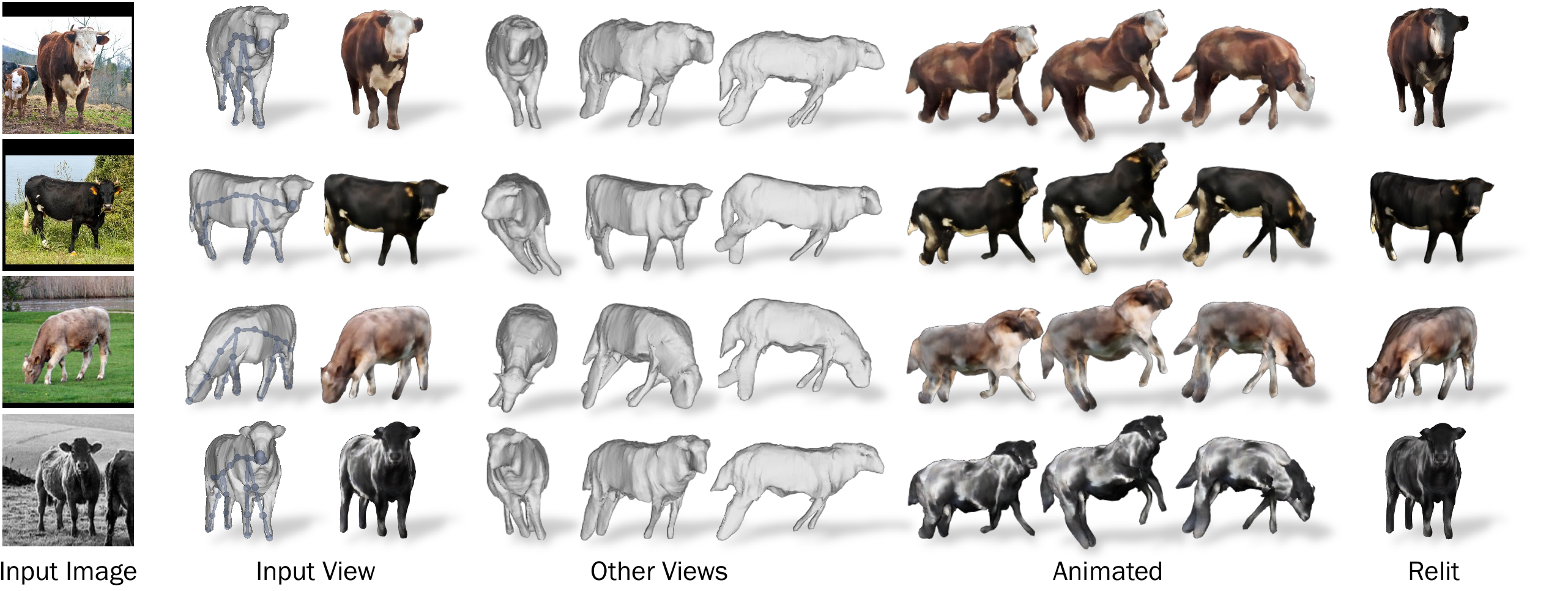}
    \caption{\textbf{Reconstruction of Giraffes, Zebras and Cows.}
    After finetuning on new categories, our method generalises to various animal classes with highly different underlying shapes.
    We show the predicted mesh from the input view and three additional views together with animated versions of the shape obtained by articulating the estimated skeleton. 
    Finally, we show a relit version.
    }
    \label{fig:supmat_recon_other_animals}
\end{figure*}

When the horse is observed from a side-view, the method might not be able to disambiguate between left and right legs, for instance, in the second to last row of \cref{fig:supmat_recon_horse_real}.
Note that our method uses only object masks and self-supervised DINO-ViT features, neither of which are sufficient to disambiguate between different legs of an animal.

\section{Additional Technical Details}
\label{sec:supmat_tech_details}
\subsection{Articulation Model Details}
Recall that our model estimates a set of bones and articulates the instance mesh using a linear blend skinning model with predicted bone rotations.
In the following, we describe in detail how the rest-pose bones are estimated and how the skinning weights are defined.

\paragraph{Bone Topology.}
\label{sec:bone_topology}
Our method only assumes a description of the topology of the animal's skeleton, and automatically estimates a set of bones at rest pose for the articulation model based on simple heuristics.
Specifically, for birds, we estimate a chain of $8$ bones with equal lengths that lie on two line segments going from the centre (root) of the rest-pose mesh to the two most extreme vertices along $z$-axis ($4$ bones on each side), forming a `spine'.

For quadrupedal animals, like horses, we lift the root joint slightly higher and further add $4$ sets of bones for modelling the legs, as illustrated in \cref{fig:supmat_compare_lassie}.
We first identify the foot joints as the lowest points of mesh (in $y$-axis) in each of four $xz$-quadrants.
We then draw $4$ line segments from the foot joints to their closest spine joints, and define a chain of $3$ bones with equal lengths on each of the segments, representing each leg.

\paragraph{Skinning Weights.}
\label{sec:skinning_weights}
Recall \ifarxiv \cref{e:skinning}, \else Eq.~(2) in the main paper, \fi where the instance mesh is further posed by a linear blend skinning equation.
Each vertex $V_{\text{ins},i}$ is associated with the bones by a skinning weight $w_{i,b}$, defined as:
\begin{equation}\label{e:skinning_w}
\begin{aligned}
  w_{i,b} &= \frac{e^{-d_{i,b}/\tau_\text{s}}}{\sum_{k=1}^{B}{e^{-d_{i,k}/\tau_\text{s}}}},
  \\
  \text{where} \quad d_{i,b} &= \min_{r \in [0,1]} \|V_{\text{ins},i} - r\tilde{\mathbf{J}}_b - (1-r)  \tilde{\mathbf{J}}_{\pi(b)}\|^2_2
\end{aligned}
\end{equation}
is the minimal distance from the vertex $V_{\text{ins},i}$ to each bone $b$, defined by the rest-pose joint locations $\tilde{\mathbf{J}}_b, \tilde{\mathbf{J}}_{\pi(b)}$ in world coordinates, and $\tau_\text{s}$ is a temperature parameter set to $0.5$.

\begin{table}[t]
\setlength{\tabcolsep}{4pt}
\centering
\caption{Ablations on Toy Bird Scans (Chamfer Distance $\downarrow$ in cm). Annotations: \faClockO\@\xspace $\tau$ schedule, \faSteam\@\xspace articulation, \faAdjust\@\xspace symmetry.}
\resizebox{0.485\textwidth}{!}{
\begin{tabular}{cccccccc}
\toprule
full     & w/o $\mathcal{R}_\text{Eik}$   & w/o $\mathcal{R}_\text{def}$   & w/o \faClockO & w/o \faAdjust & w/o \faSteam \ \faAdjust  & w/o $\mathcal{L}_{\text{im}}$  & 4 / 16 bones  \\ 
\midrule
\textbf{0.79}  & 1.19  &  1.37  &  1.66  &  1.56  &  1.63  &  2.11  & 2.10 / 1.27   \\
\bottomrule
\end{tabular}
}
\label{tab:supmat_ablation}
\end{table}

\paragraph{Constraints on the Bone Rotations.}
Our model learns complex articulated 3D poses of animals using reconstruction losses on single-view images, without any explicit 3D geometric supervision, which is an extremely ill-posed task.
In order to prevent unnatural poses, we enforce minimal constraints on the bone rotations: (1) all bone rotations are limited to $(-60^\circ, 60^\circ)$, (2) for quadrupeds, leg rotations around $y$- and $z$-axes (`twist' and `side-bending') are further limited to $(-18^\circ, 18^\circ)$.

\subsection{Network Architectures}
We implement the feature field $\psi$, template SDF $s$ and the light network $f_{\text{l}}$ using 5-layer MLPs, and the albedo field $f_{\text{a}}$ and deformation field $f_{\Delta V}$ with 8-layer MLPs.
The articulation network consists of $4$ transformer blocks.
All coordinate inputs are encoded using $\texttt{sin}(\cdot)$ and  $\texttt{cos}(\cdot)$ with $8$ frequencies.

The encoders are simple convolutional networks, described in \cref{tab:arch_encoder}.
In practice, the viewpoint network $f_\text{vp}$ is also (separately) implemented using the same architecture as the encoder.
Abbreviations of the components are defined as follows:
\begin{itemize}
	\item $\text{Conv}(c_{in}, c_{out}, k, s, p)$: 2D convolution with $c_{in}$ input channels, $c_{out}$ output channels, kernel size $k$, stride $s$ and padding $p$
	\item $\text{GN}(n)$: group normalization~\cite{wu2018group} with $n$ groups
	\item $\text{LReLU}(p)$: leaky ReLU~\cite{maas2013rectifier} with a slope $p$
\end{itemize}

\subsection{Hyper-parameters and Training Details}
All hyper-parameters are listed in \cref{tab:params}.
We enable the articulation after $10$k iterations and the deformation after $40$k iterations, to prevent the model from overfitting individual images with excessive articulation and deformation.
During the first $5$k iterations, we allow the model to explore all four viewpoint hypotheses by randomly sampling the four hypotheses uniformly, and gradually decrease the chance of random sampling to $20\%$ while sampling the best hypothesis for the rest $80\%$ of the time.
The temperature $\tau$ is decreased from $1$ to $0.01$ over the course of $100$k iterations.
It takes roughly $20$ hours to train the full model for $150$k iterations on one single NVIDIA A40 GPU.

\subsection{Keypoint Transfer Evaluation Details}
Due to the lack of 3D ground-truth for in-the-wild objects, we employ the Keypoint Transfer task and compute the Percentage of Correct Keypoints (PCK)~\cite{kanazawa18cmr, Li2020umr, kulkarni20articulation-aware} as an indirect metric for evaluating the reconstructed 3D shapes, as described in \ifarxiv \cref{sec:compare}. \else Sec.~4.4 in the main paper. \fi
A transferred keypoint is correct if it is within a distance $d$ of the corresponding ground-truth 2D keypoint in the target image. The value of $d$ is computed as $0.1 \cdot \max(h, w)$ for PCK@0.1, where $h$ and $w$ represent the height and width of the ground-truth object bounding box.
We follow the open-source implementation\footnote{\cite{Li2020umr}: \url{https://github.com/NVlabs/UMR} ,\cite{kulkarni20articulation-aware}: \url{https://github.com/nileshkulkarni/acsm/}} of the metric as described in \cite{kulkarni20articulation-aware} for the PASCAL VOC Horse dataset and in \cite{Li2020umr} for the CUB Bird dataset.
It should be noted that \cite{kulkarni20articulation-aware} defines their error with respect to a bounding box that is padded by $5\%$ of the original size on each side. To maintain consistency, we follow the same practice for the PASCAL VOC Horse dataset.

\begin{table}[t]
\small
\begin{center}
\caption{Architecture of the patch feature encoders $f_\text{k}$, $f_\text{o}$.}
\begin{tabular}{lc}
\toprule
 Encoder & Output size \\ \midrule
 Conv(384, 256, 4, 2, 1) + GN(64) + LReLU(0.2) & 16 $\times$ 16\\
 Conv(256, 256, 4, 2, 1) + GN(64) + LReLU(0.2) & 8 $\times$ 8\\
 Conv(256, 256, 4, 2, 1) + GN(64) + LReLU(0.2) & 4 $\times$ 4\\
 Conv(256, 256, 4, 2, 0) $\rightarrow$ output  & 1 $\times$ 1\\
\bottomrule
\end{tabular}
\end{center}
\label{tab:arch_encoder}
\end{table}
\begin{table}[t]
\small
\begin{center}
\caption{Training details and hyper-parameter settings.}\label{tab:params}
\begin{tabular}{lc}
\toprule
 Parameter & Value/Range \\ \midrule
 Optimiser & Adam \\
 Learning rate on prior ($\psi$ and $s$) & $1\times 10^{-3}$ \\
 Learning rate on others & $1\times 10^{-4}$ \\
 Number of iterations & $150$k \\
 Batch size & $10$ \\
 Loss weight $\lambda_{\text{m}}$ & $10$ \\
 Loss weight $\lambda_{\text{dt}}$ & $10$ \\
 Loss weight $\lambda_{\text{im}}$ & $1$ \\
 Loss weight $\lambda_{\text{f}}$ & $10$ \\
 Loss weight $\lambda_{\text{E}}$ & $0.01$ \\
 Loss weight $\lambda_{\text{d}}$ & $10$ \\
 Loss weight $\lambda_{\text{a}}$ & $0.1$ \\
 Loss weight $\lambda_{\text{h}}$ & $1$ \\ \midrule
 Image size & $256 \times 256$ \\
 Field of view (FOV) & $25^\circ$ \\
 Camera location & $(0, 0, 10)$ \\
 Tetrahedral grid size & $256$ \\
 Initial mesh centre & $(0, 0, 0)$ \\
 Translation in $x$- and $y$-axes & $(-0.4, 0.4)$ \\
 Translation in $z$-axis & $(-1.0, 1.0)$ \\
 Number of spine bones & $8$ \\
 Number of bones for each leg & $3$ \\
 Viewpoint hypothesis temperature $\tau$ & $(0.01, 1.0)$ \\
 Skinning weight temperature $\tau_\text{s}$ & $0.5$ \\
 Ambient light intensity $k_s$ & $(0.0, 1.0)$ \\
 Diffuse light intensity $k_d$ & $(0.5, 1.0)$ \\
\bottomrule
\end{tabular}
\end{center}
\end{table}

\end{document}